\documentclass[11pt]{article}
\usepackage{acl}
\usepackage{times}
\usepackage{latexsym}
\usepackage{algorithm}
\usepackage{algorithmic}
\usepackage[T1]{fontenc}

\usepackage[utf8]{inputenc}

\usepackage{microtype}
\usepackage{graphicx}
\usepackage{subcaption}
\usepackage{booktabs} 
\usepackage{inconsolata}

\usepackage{hyperref}

\usepackage{amsmath}
\usepackage{amssymb}
\usepackage{mathtools}
\usepackage{amsthm}

\usepackage[capitalize,noabbrev]{cleveref}

\theoremstyle{plain}
\newtheorem{theorem}{Theorem}[section]

\newtheorem{corollary}[theorem]{Corollary}
\theoremstyle{definition}

\theoremstyle{remark}

\usepackage[textsize=tiny]{todonotes}

\usepackage{amsmath,amsfonts,bm}

\def\eqref#1{equation~\ref{#1}}

\def\1{\bm{1}}

\DeclareMathAlphabet{\mathsfit}{\encodingdefault}{\sfdefault}{m}{sl}
\SetMathAlphabet{\mathsfit}{bold}{\encodingdefault}{\sfdefault}{bx}{n}

\usepackage{url}

\newcommand{\RETURN}{\STATE \textbf{return} }
\usepackage{wrapfig}
\usepackage{lineno}
\usepackage{multirow}
\usepackage{comment}
\usepackage{tikz}
\usetikzlibrary{calc}
\usepackage{makecell}
\usepackage{xcolor}
\usepackage{colortbl}
\usepackage{caption}
\usepackage{pifont} 
\usepackage{fontawesome}
\usepackage{xspace}
\usepackage[english]{babel}
\usepackage{tcolorbox}
\usepackage{enumitem}

\definecolor{lightblue}{rgb}{0.85, 0.92, 1.0}
\definecolor{lightred}{rgb}{1.0, 0.9, 0.9}

\newcommand{\rui}[1]{\textbf{\textcolor{brown}{}}}
\newcommand{\soumik}[1]{\textbf{\textcolor{brown}{}}}

\newcommand{\ashirbad}[1]{\textbf{\textcolor{brown}{}}}

\newlength{\barunit}
\newcommand{\oursb}{\textsc{EqSpec}\xspace}
\newcommand{\oursx}{\textsc{EXSpec}\xspace}

\begin{document}

  \title{Correctness Forensics for Batch Speculative Decoding: Diagnosing the Ragged Tensor Problem\thanks{Accepted to Findings of the Association for Computational Linguistics: EMNLP 2026.}}
    \author{First Author \\
      Affiliation / Address line 1 \\
      Affiliation / Address line 2 \\
      Affiliation / Address line 3 \\
      \texttt{email@domain} \\\And
      Second Author \\
      Affiliation / Address line 1 \\
      Affiliation / Address line 2 \\
      Affiliation / Address line 3 \\
      \texttt{email@domain} \\}



\author{
  \textbf{Ranran Haoran Zhang\textsuperscript{1}},
  \textbf{Soumik Dey\textsuperscript{2}},
  \textbf{Ashirbad Mishra\textsuperscript{2}},\\
  \textbf{Hansi Wu\textsuperscript{2}},
  \textbf{Binbin Li\textsuperscript{2}},
  \textbf{Rui Zhang\textsuperscript{1}}\\
  \textsuperscript{1}The Pennsylvania State University,
  University Park, PA, USA\\
  \textsuperscript{2}eBay Inc., San Jose, CA, USA\\
  {\small\texttt{\{hzz5361, rmz5227\}@psu.edu}}\\
  {\small
    \textbf{Code:}
    \href{https://github.com/eBay/spec_dec}{\texttt{https://github.com/eBay/spec\_dec}}
  }
}





\maketitle

\begin{abstract}
Inference optimizations are routinely evaluated by throughput alone, without verifying output correctness. We conduct a forensic analysis of batch speculative decoding and find that several widely-used implementations silently produce corrupted outputs (repetitive tokens, \texttt{<unk>} symbols) while reporting competitive speed; failures invisible to metrics like ROUGE. We trace the root cause to the ragged tensor problem: variable token acceptance desynchronizes position IDs, attention masks, and KV-cache across a batch. We formalize the synchronization invariants (rectangular alignment and position/KV-cache correspondence) that valid contiguous batching must preserve and show that maintaining them incurs superlinear alignment overhead. \oursb enforces the invariants without custom kernels; \oursx schedules same-length sequences to bypass realignment. On SpecBench across three model families, \oursx reaches $3\times$ throughput at batch size 8. Our methods remove catastrophic synchronization failures but achieve 90.8--97.3\% exact match against batch-size-1 non-speculative decoding. The residual mismatches are consistent with batch-dependent floating-point effects.
\end{abstract}

\section{Introduction}

LLM inference optimizations are routinely evaluated by throughput alone, with output correctness assumed rather than verified. We argue that \emph{output equivalence}, the requirement that an optimized system reproduce the same distribution as standard decoding, should be a first-class evaluation criterion. A system that produces garbage quickly is not an improvement.

This paper conducts a forensic analysis of batch speculative decoding using an observe$\rightarrow$diagnose$\rightarrow$formalize$\rightarrow$validate audit protocol. Speculative decoding~\citep{leviathan2023fast,chen2023accelerating} has a small draft model propose tokens that a target model verifies in parallel; its defining requirement is output equivalence~\citep{leviathan2023fast,xia-etal-2024-unlocking}. Extending speculation to batches raises the ragged tensor problem: sequences accept different numbers of draft tokens, desynchronizing position IDs, attention masks, and KV-cache on the standard contiguous layout.

Figure~\ref{fig:yesbut} shows that several widely-used implementations, BSP~\citep{su2023synergy} and DSD~\citep{yan2025decoding}, report competitive throughput while producing repetitive tokens or \texttt{<unk>} symbols under greedy decoding. These are fundamental correctness violations that metrics such as ROUGE~\citep{qian2024bass} can mask, raising a \textbf{reproducibility concern}: published speedups may not be comparable unless output equivalence is verified. We use exact match and partial match (fraction of tokens before first divergence) to detect corruption and localize its cause: immediate divergence indicates position-ID errors; gradual decay reveals KV-cache drift.

\definecolor{spec1gray}{HTML}{808080}
\definecolor{dsdcoral}{HTML}{ff6b6b}
\definecolor{bspdarkred}{HTML}{8b0000}
\definecolor{oursdarkblue}{HTML}{1e3a8a}
\definecolor{oursskyblue}{HTML}{60a5fa}

\begin{figure*}[t]
    \centering
    
    \begin{minipage}[t]{0.4\textwidth}
        \centering
        \tikz[baseline=(current bounding box.north)]{
            \node[fill=green!20, rounded corners, inner sep=8pt, font=\large\bfseries] {YES, High Throughput...};
        }

    \includegraphics[width=0.95\linewidth]{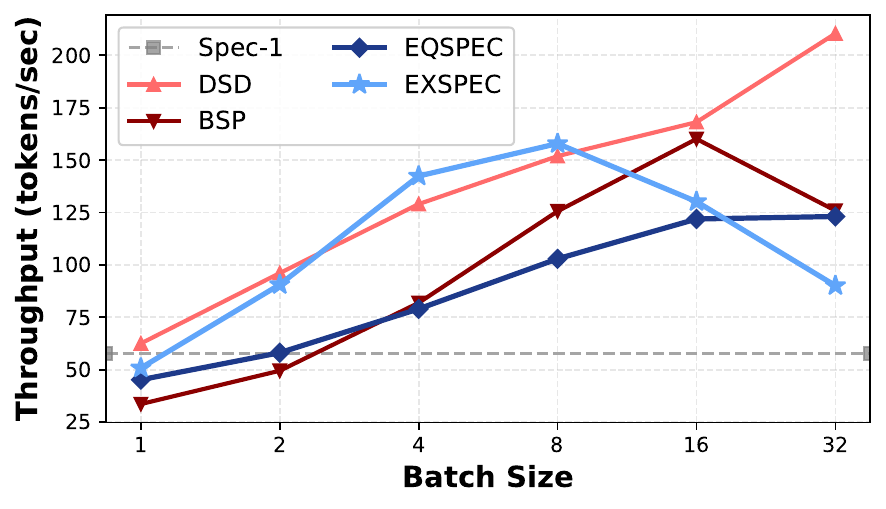}
        
        
        

    \end{minipage}
    \hfill
    \begin{minipage}[t]{0.02\textwidth}
        \centering
        \tikz{
            \draw[->, ultra thick, red!70, line width=3pt] (0,0) -- (0.5,0);
        }
    \end{minipage}
    \hfill
    \begin{minipage}[t]{0.42\textwidth}
        \centering
        \tikz[baseline=(current bounding box.north)]{
            \node[fill=red!20, rounded corners, inner sep=8pt, font=\large\bfseries] {BUT, Output Gibberish!};
        }
        \vspace{3mm}
        
        \scriptsize
        \setlength{\tabcolsep}{4pt}
\begin{tabular}{lp{3.9cm}}
\toprule
\textbf{Method} & \textbf{Generated Output} \\
\midrule
\multicolumn{2}{l}{\textit{Prompt: "The weather today is"}} \\
\midrule
Ground Truth & \textcolor{black}{not as bad as it was yesterday...} \\
\midrule
\rowcolor{spec1gray!15}
Spec-1 & \textcolor{green!70!black}{\ding{51}~Matches ground truth} \\
\rowcolor{bspdarkred!20}
BSP & \textcolor{red}{\ding{55}~\textbf{not not not not not not not...}} \\
\rowcolor{dsdcoral!25}
DSD & \textcolor{red}{\ding{55}~\textbf{\textlangle unk\textrangle\textlangle unk\textrangle~The perfect time...}} \\
\midrule
\rowcolor{oursdarkblue!15}
\oursb & \textcolor{green!70!black}{\ding{51}~Matches ground truth} \\
\rowcolor{oursskyblue!20}
\oursx & \textcolor{green!70!black}{\ding{51}~Matches ground truth} \\
\bottomrule
\end{tabular}
        
        \vspace{-2.5mm}

    \end{minipage}
    

    \caption{Batch speculative decoding on Vicuna-7B/68M: Existing methods  achieve high throughput but \textbf{violate the fundamental requirement of output equivalence} by producing corrupted outputs. \oursb and \oursx avoid this catastrophic corruption while retaining competitive performance.
    }
    \label{fig:yesbut}
    \vspace{-15pt}
\end{figure*}

We trace these failures to the ragged tensor problem~\citep{qian2024bass}. For contiguous layouts, we formalize rectangular alignment~(I1) and position/KV-cache correspondence~(I2) as synchronization conditions for algorithmic output equivalence~(Section~\ref{sec:dynamic-padding}). \oursb enforces them without custom kernels. The invariants require realigning the batch after every verification round; this overhead grows superlinearly, reaching 40\% at batch size 8~(Section~\ref{sec:algo_analysis}), a cost intrinsic to contiguous layouts that explains why prior methods either break equivalence or fail to scale. \oursx groups same-length sequences to bypass realignment, reaching $3\times$ throughput at batch size 8~(Section~\ref{sec:overhead-and-scaling}).

We instantiate this audit for batch speculative decoding: task-specific diagnostics, synchronization invariants tied to the ragged-tensor problem, and empirical validation. The invariants concern properties of tensor-batched inference (position tracking, mask alignment, cache synchronization) that arise whenever sequences grow at different rates; we validate them only for batch speculative decoding. Our contributions:
\begin{itemize}
    \item \textbf{A forensic audit} revealing that several batch speculative decoding implementations produce corrupted outputs invisible to standard metrics, with diagnostics that distinguish failure signatures and isolate root causes~(Appendix~\ref{app:bug_taxonomy}).
    \item \textbf{A formal correctness specification:} two synchronization invariants for output equivalence, enforced by \oursb without custom kernels. Across three model families, our methods achieve 90.8--97.3\% exact match against batch-size-1 non-speculative decoding.
    \item \textbf{A cost analysis and scheduling-based mitigation:} realignment overhead grows superlinearly under contiguous layouts; \oursx groups same-length sequences to bypass it, reaching $3\times$ throughput at batch size 8.
\end{itemize}

\section{Failure Modes and the Audit Protocol}
\label{sec:design-space}

\begin{figure*}
    \vspace{-10pt}
    \centering
    \includegraphics[width=0.75\linewidth]{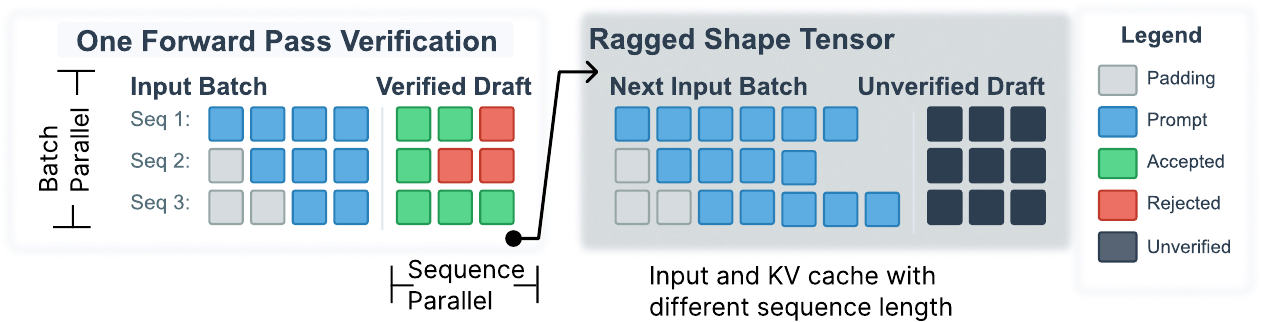}
    \caption{The ragged tensor problem in batch speculative decoding. 
    Differing numbers of accepted draft tokens across sequences in the same batch lead to ragged-shaped input IDs tensors and KV-cache that disrupt subsequent batch operations.
    }
    \label{fig:intro}
\end{figure*}

Inference optimizations that modify batch-level tensor state (position IDs, attention masks, KV-cache) are prone to silent correctness failures. We organize our analysis as a four-stage audit: \emph{observe} divergence from a reference oracle under deterministic decoding; \emph{diagnose} via exact and partial match, separating immediate position-ID failures from gradual KV-cache drift; \emph{formalize} the invariants each failure violates (Section~\ref{sec:dynamic-padding}); \emph{validate} that enforcing them restores equivalence and quantify the cost (Section~\ref{sec:exp}).

We apply this to batch speculative decoding. When sequences accept different numbers of draft tokens, the resulting tensors become irregular: the ragged-tensor problem (Figure~\ref{fig:intro}). Approaches divide by \emph{KV-cache layout} (Table~\ref{tab:design-space}). \textbf{Contiguous} layouts (Family~A) maintain a rectangular $[\text{batch},\text{seq}]$ tensor and must restore that shape after every verification round. \textbf{Variable-length} layouts (Family~B) pack sequences into a flat buffer indexed by cumulative lengths, dissolving the alignment requirement by construction; this is the route taken by production systems (vLLM~v1, SGLang).

\begin{table*}[t]
\centering
\footnotesize
\setlength{\tabcolsep}{4pt}
\begin{tabular}{llcccc}
\toprule
\textbf{Family} & \textbf{Strategy} & \textbf{Correct} & \textbf{Keeps all} & \textbf{Portable} & \textbf{Bounded mem.} \\
 & & \textbf{\footnotesize(output-equiv.)} & \textbf{\footnotesize tokens} & \textbf{\footnotesize(std.\ kernels)} & \textbf{\footnotesize(no $K\times$)} \\
\midrule
\multirow{4}{*}{\textbf{A}: contiguous}
 & Masking                          & \ding{55} & \ding{51} & \ding{55} & \ding{51} \\
 & Rollback                         & \ding{51} & \ding{55} & \ding{51} & \ding{51} \\
 & Batch expansion                  & \ding{51} & \ding{51} & \ding{51} & \ding{55} \\
 & \textbf{Dynamic padding (ours)}  & \ding{51} & \ding{51} & \ding{51} & \ding{51} \\
\midrule
\textbf{B}: variable-length & Flat / paged packing & \ding{51} & \ding{51} & \ding{55} & \ding{51} \\
\bottomrule
\end{tabular}
\caption{Design space for the ragged-tensor problem, organized by KV-cache memory layout.}
\label{tab:design-space}
\end{table*}

\paragraph{Family~A: contiguous layout.} Four strategies attempt to keep the rectangular tensor, each producing a distinct failure signature.
\emph{Masking} reassigns position IDs for rejected tokens but leaves the KV-cache desynchronized; BSP~\citep{su2023synergy} and EAGLE's experimental batch code\footnote{\url{https://github.com/SafeAILab/EAGLE/issues/250}}~\citep{li2025eagle3scalinginferenceacceleration} take this route and corrupt output (Figure~\ref{fig:yesbut}). Our partial-match diagnostic reveals BSP's signature: initially correct output before KV-cache drift triggers repetitive loops (gradual decay, not immediate failure). Making masking correct requires custom kernels~\citep{qian2024bass}.
\emph{Rollback} truncates all sequences to the batch-minimum accepted length~\citep{wolf-etal-2020-transformers,rollback}: correct but discards verified tokens, with waste growing until throughput collapses.
\emph{Batch expansion} (vLLM~v0) duplicates each sequence into $K$ draft prefixes: correct but $K\times$ memory, deprecated due to OOM failures.
\emph{Dynamic padding} realigns the batch via unpad--append--repad each round; DSD~\citep{yan2025decoding} attempts it but desynchronizes position IDs/KV-cache and samples the bonus token from the draft model. Our diagnostic shows near-zero partial match, indicating immediate divergence.
Only dynamic padding is correct, token-preserving, portable, and memory-bounded, provided synchronization is done correctly, which none of the implementations we tested achieves (Appendix~\ref{app:bug_taxonomy}). 




\paragraph{Family~B: variable-length layout.} Continuous-batching systems (vLLM~v1, SGLang) store the KV-cache in a flat buffer indexed by cumulative lengths, so ragged acceptance never requires repadding and I1 does not arise. Position consistency (I2) is still required: rejected tokens must be rolled back in both position IDs and the KV-cache. This layout depends on paged-attention kernels available only on CUDA/Triton; on other accelerators (Ascend, Intel Gaudi, Apple Metal) and for offline batch processing, the contiguous family remains the only option. Family~B is thus the production default where its kernels exist; the contiguous family, our focus, covers the remaining hardware and workloads (Appendix~\ref{app:vllm_sglang}).

\begin{figure}[ht]
    \centering
    \includegraphics[width=0.7\linewidth]{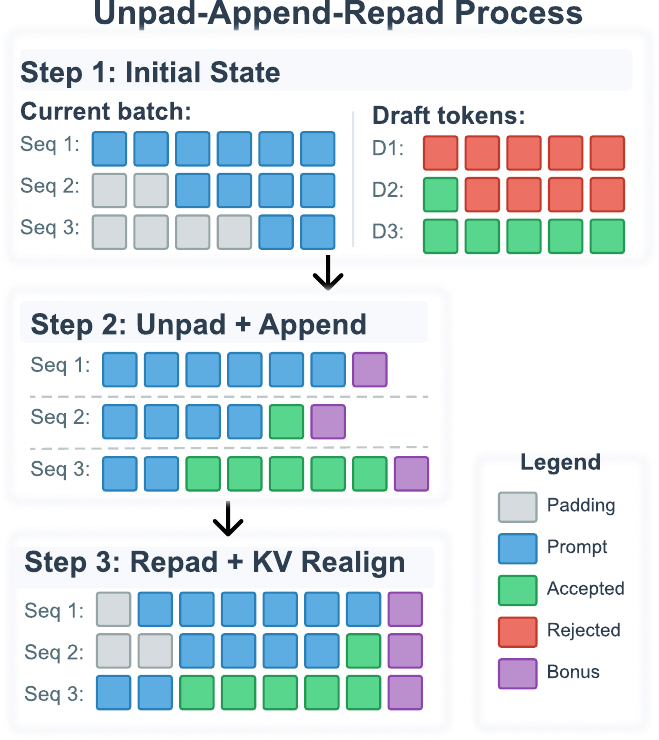}
\caption{\oursb synchronizes via unpad--append--repad.}
\label{fig:unpad-append-repad}
\end{figure}

\subsection{Output Equivalence as a Diagnostic Criterion}
\label{sec:output-equiv}

At the algorithmic level, \emph{output equivalence} requires greedy decoding to produce the same token sequence as standard autoregressive generation; for temperature${}>0$, it requires preserving the target distribution. Batched floating-point execution can alter argmax decisions, so empirical exact match diagnoses corruption but does not guarantee bit-exact identity. We use greedy decoding because it exposes synchronization errors more clearly than stochastic sampling. BSP and DSD fail this diagnostic regardless of throughput (Figure~\ref{fig:yesbut}).

\section{Correctness Specification and Cost}
\label{sec:method}

Having classified failure modes (Stage 1--2), we now formalize the invariants correctness requires and quantify their cost (Stage 3--4). \oursb specifies and enforces the minimal synchronization invariants (Section~\ref{sec:dynamic-padding}); \oursx reduces their overhead by grouping same-length sequences for which realignment vanishes (Section~\ref{sec:algo_analysis}).


\subsection{\oursb: Formalizing the Correctness Specification (Stage 3)}
\label{sec:dynamic-padding}

The diagnostic stage identified \emph{what} fails; we now formalize \emph{why}---the invariants whose violation produces each observed failure signature.

\paragraph{Rectangular Alignment.} Let $B$ denote the batch size, and let sequence $i$ have padding length $p_i$, content length $c_i$, and $a_i$ newly accepted tokens (including the bonus token) after verification round $t$. A valid implementation must maintain: 
\begin{equation*} 
\tag{\textbf{I1}} \label{eq:rect_align}
\quad \forall i \in [1,B],\quad  p_i^{(t)} + c_i^{(t)} = L^{(t)} 
\end{equation*}
where $L^{(t)}$ is the padded sequence length at round $t$. 

\paragraph{Position and KV-Cache Correspondence.} Position IDs must be derived consistently from the attention mask: \begin{equation*}
\tag{\textbf{I2}} \label{eq:position_id}
\quad \mathrm{pos}_i[j] = \max\left(0, \sum_{k=0}^{j} \mathbf{1}[\mathrm{tok}_i[k] \neq \texttt{pad}] - 1\right) \end{equation*}
which guarantees contiguous position IDs starting from zero at the first content token, as required by RoPE~\citep{su2024roformer} and other position-aware attention mechanisms. At each content position, the KV-cache entry must also be the target-computed entry for the token at that logical position.
Continuous-batching systems (vLLM, SGLang) sidestep I1 through variable-length packing (Family~B, Section~\ref{sec:design-space}), but I2 remains a prerequisite for correctness regardless of batching strategy. 

After batch verification, sequence $i$ accepts $a_i$ tokens, breaking rectangular alignment since $a_i$ varies across sequences. To restore I1, the new padding must satisfy: 
\begin{equation} \forall i \in [1,B], \quad p_i^{(t+1)} + c_i^{(t)} + a_i = L^{(t+1)}
\end{equation} 
where $L^{(t+1)} = \max_j(c_j^{(t)} + a_j)$. 
This yields the per-sequence padding offset: 
\begin{equation} \delta_i = p_i^{(t+1)} - p_i^{(t)} = \left(L^{(t+1)} - L^{(t)}\right) - a_i \end{equation}
We prove that this update rule preserves rectangular alignment (I1) and the cache-correspondence clause of I2 across all verification rounds (Theorem~\ref{theorem}): because realignment relocates cached entries rather than recomputing them, it introduces no numerical drift. The position-ID clause of I2 is enforced by recomputing position IDs after each repadding operation. Together with target-model bonus sampling, maintaining I1 and I2 is sufficient for algorithmic output equivalence (Corollary~\ref{cor:equiv}).


Algorithm~\ref{alg:batch_speculative} summarizes \oursb's three-phase loop.


\begin{algorithm}[t]
\footnotesize
\caption{\oursb: Minimalism Batch Spec}
\label{alg:batch_speculative}
\begin{algorithmic}[1]
\REQUIRE Draft model $\mathcal{M}_d$, Target model $\mathcal{M}_t$, Prompts $\mathcal{P}$, Max tokens $T$, Draft length $K$
\ENSURE Generated sequences $\mathcal{S}$
\STATE $\mathcal{S}, L \gets \texttt{Tokenize}(\mathcal{P})$ \COMMENT{Left-pad batch to width $L$}
\STATE $\textit{KVCache} \gets \emptyset$
\WHILE{until max new tokens}
    \STATE \textbf{Phase 1: Draft Generation}
    \STATE $D \gets \mathcal{M}_d.\texttt{Generate}(\mathcal{S}, K)$
    \STATE \textbf{Phase 2: Batch Verification}
    \STATE \COMMENT{See Alg. 3 in Appendix for BatchVerify}
    \STATE $A, B, \textit{KVCache} \gets \texttt{BatchVerify}(\mathcal{M}_t, \mathcal{S}, D, \textit{KVCache})$
    \STATE \COMMENT{See Figure~\ref{fig:unpad-append-repad} for illustration on index offset.}

    \STATE \textbf{Phase 3: Realign by Offset $\delta$}
    \FOR{each sequence $i$ in batch}
        \STATE $\mathcal{S}[i] \gets \mathcal{S}[i] \oplus A[i] \oplus B[i]$ \COMMENT{Append accepted + bonus; $a_i\!=\!|A[i]|\!+\!1$}
    \ENDFOR
    \STATE $L' \gets \max_j (c_j + a_j)$ \COMMENT{New batch width $L^{(t+1)}$}
    \STATE $\delta_i \gets (L' - L) - a_i \quad \forall i$ \COMMENT{Per-sequence offset, Eq.~2}
    \STATE $\mathcal{S}, \textit{KVCache}, L \gets \texttt{Realign}(\mathcal{S}, \textit{KVCache}, \delta)$ \COMMENT{Shift padding \& KV by $\delta_i$; reset pos-IDs (I2)}
\ENDWHILE
\RETURN $\mathcal{S}$
\end{algorithmic}%
\end{algorithm}

\paragraph{Unpad-Append-Repad.}
Figure~\ref{fig:unpad-append-repad} illustrates the realignment phase. The offset $\delta_i$ realigns all batch state: token sequences receive fresh left padding, attention masks are recomputed, position IDs are recalculated via I2, and KV-cache entries are shifted by $\delta_i$ positions. The bonus token, sampled from the target distribution after the forward pass completes (Algorithm~3, Appendix~\ref{app:batchverify}), lacks KV-cache entries and must be included in the next forward pass.

\begin{figure}
    \centering
    \includegraphics[width=0.7\linewidth]{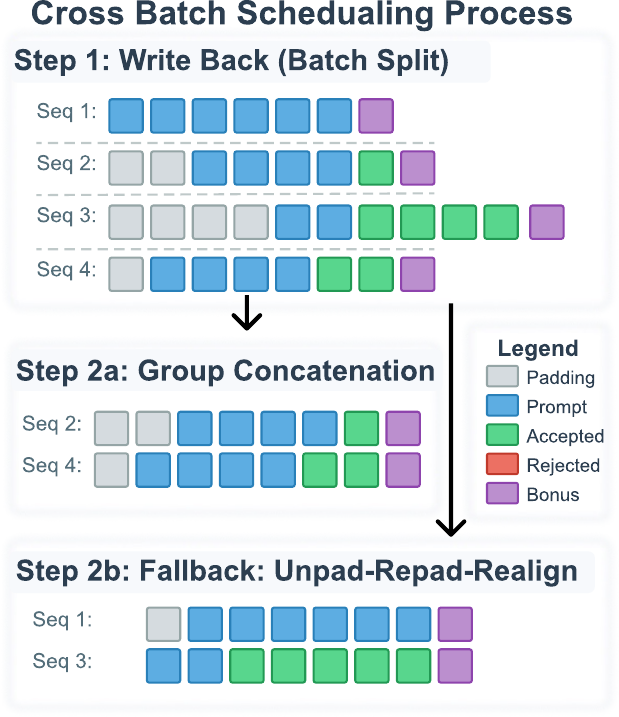}
    \caption{\oursx pools ragged sequences by length, avoiding realignment; only unmatched sequences need syncing, turning fixed overhead into optional cost.}
    \label{fig:xbatch}
\end{figure}

\subsection{\oursx: Quantifying and Reducing the Cost of Correctness (Stage 3--4)}
\label{sec:algo_analysis}
\paragraph{Alignment Overhead.} Beyond the drafting and verification costs ($c_{\text{draft}}$, $c_{\text{verify}}$) of single-sequence speculation~\citep{leviathan2023fast}, batched speculation adds an alignment overhead $c_{\text{overhead}}(B) = c_{\text{pad}}(B) + c_{\text{kv}}(B)$, absent from single-sequence analysis, where $c_{\text{pad}}(B)$ covers unpad--append--repad and $c_{\text{kv}}(B)$ KV-cache realignment. The rest of this section quantifies it.

\begin{theorem}[Alignment Overhead under Contiguous Layout]
\label{thm:superlinear}
Restoring rectangular alignment (Invariant~I1) each round pads every sequence up to the batch's longest acceptance $M_B=\max_i a_i$, so the per-round realignment work is the total gap $\sum_{i=1}^{B}(M_B-a_i)$. With acceptances $a_i\in[0,K]$ drawn i.i.d.\ with mean $\mu$, where $K$ is the per-round draft budget, this work has expectation $B\,(E[M_B]-\mu)$. Since $E[M_B]$ rises with $B$, it grows strictly faster than linearly over the batch sizes of interest, reaching the linear rate $(K-\mu)B$ only as $E[M_B]\to K$. The alignment cost $c_{\text{overhead}}(B)$ scales with this work.
\end{theorem}

This cost is a property of the contiguous layout that variable-length layouts (Family~B) avoid; within it, the KV-cache dominates, since its rank-4 tensors (batch $\times$ heads $\times$ sequence $\times$ dimension) are reallocated on every padding change while $c_{\text{draft}}$ and $c_{\text{verify}}$ stay stable under GPU parallelism. The alignment overhead also grows with variance in acceptances across sequences; Section~\ref{sec:overhead-and-scaling} confirms this empirically.
This explains the pattern the audit's diagnostic stage surfaced: prior methods either skip synchronization (breaking equivalence) or pay the full superlinear cost (failing to scale)---no middle ground under contiguous layouts.

However, the speedup analysis reveals a key insight: when all sequences accept the same number of tokens, $\delta_i = 0$ and realignment overhead vanishes (Corollary~\ref{corollary_xbatch}). Rather than accelerating these operations, \oursx sidesteps them through scheduling.
\paragraph{Cross-Batch Scheduling.}\oursx differs from \oursb in how it manages sequence lifecycles. Rather than maintaining fixed batches that require realignment after every verification, we introduce a \emph{SequencePool} that holds sequences individually in their ragged states. This enables three optimizations: (i) sequences that complete (reach EOS) are immediately removed, avoiding wasted computation; (ii) \emph{lazy realignment} defers synchronization until strictly necessary; and (iii) \emph{dynamic batch formation} over a sliding window of $W > B$ sequences greatly increases the chance of finding same-length groups.

Figure~\ref{fig:xbatch} illustrates the flow. After verification, ragged sequences return directly to the pool without realignment. The scheduler scans the active window and attempts to form batches of identical length. Same-length sequences concatenate directly (no padding adjustments, no position-ID recomputation, no KV-cache realignment), bypassing $c_{\text{overhead}}(B)$ entirely. Only when same-length grouping fails do we fall back to unpad--append--repad. Combined with prompt-length sorting, grouping rates approach unity for similar workloads, turning constant realignment into the rare case. The complete procedure is detailed in Algorithm~\ref{alg:cross_batch_speculative} (Appendix~\ref{app:exspec}).

\section{Experimental Validation}
\label{sec:exp}
We validate the diagnostics and the correctness specification along two dimensions conflated in prior work: \textbf{output equivalence} and \textbf{throughput}.

\subsection{Experimental Setup}

\paragraph{Models.}
To test robustness across architectures, we evaluate three target--draft pairs: Vicuna-7B/68M~\citep{zheng2023judging}, Qwen3-8B/0.6B~\citep{qwen3}, and GLM-4-9B/0.6B~\citep{glm2024chatglm}. Unless otherwise noted, experiments use NVIDIA A100 80GB GPUs, PyTorch 2.7, HuggingFace Transformers 4.51.3, five draft tokens per speculation round, and greedy decoding for determinism.

\paragraph{Evaluation and Datasets.}
We use SpecBench~\citep{xia-etal-2024-unlocking}, a widely adopted benchmark for speculative decoding research. We also use Multi30k~\citep{multi30k} for a controlled \oursx study that contrasts random sampling with an identical-length subset, isolating sequence-length diversity as the driver of grouping rate. For the main evaluation, we measure: (1) \textit{Throughput}: tokens/s across batch sizes; and (2) \textit{Output Equivalence}: exact-match rate (full-sequence equivalence with non-speculative decoding) and partial-match rate (fraction of tokens matching until the first divergence). The partial-match metric helps localize failure modes. Early divergence typically indicates position-ID or KV-cache misalignment.


\paragraph{Batch Speculative Decoding Compared.}
Following our failure-mode taxonomy (Section~\ref{sec:design-space}), we evaluate:
(1) \textit{Masking approaches}: \textbf{BSP}~\citep{su2023synergy} attempts masking with adaptive speculation but suffers position-ID inconsistencies (BASS~\citep{qian2024bass} also follows this approach but requires custom CUDA kernels, limiting generality);
(2) \textit{Dynamic-padding approaches}: \textbf{DSD}~\citep{yan2025decoding} explores padding but mishandles the KV-cache, while \textbf{\oursb} implements correct synchronization and \textbf{\oursx} adds cross-batch scheduling;
(3) \textit{Reference baselines}: \textbf{Spec-1} (batch-size-1 speculation from Hugging Face Transformers), which does not support batch speculative decoding\footnote{\url{https://github.com/huggingface/transformers/issues/32165}}.
Production systems (vLLM, SGLang; see also Sections~\ref{sec:design-space} and~\ref{sec:dynamic-padding}) employ continuous batching with paged attention, which sidesteps the ragged tensor problem through different architectural choices. SGLang supports only EAGLE-family drafters; vLLM~v1 serves as a numerical control in Section~\ref{sec:verification_exp}, while broader vLLM and SGLang results remain in Appendix~\ref{app:vllm_sglang}. No existing implementation uses rollback due to its inherent wastefulness.


\begin{table*}[t]
\centering
\footnotesize
\setlength{\tabcolsep}{4pt}
\begin{tabular}{l|rr|rr|rr|rr|rr|rr}
\toprule
\multirow{3}{*}{\textbf{Method}} & \multicolumn{4}{c|}{\textbf{Vicuna}} & \multicolumn{4}{c|}{\textbf{Qwen3}} & \multicolumn{4}{c}{\textbf{GLM4}} \\
& \multicolumn{2}{c|}{\textbf{Batch 1}} & \multicolumn{2}{c|}{\textbf{Batch 4}} & \multicolumn{2}{c|}{\textbf{Batch 1}} & \multicolumn{2}{c|}{\textbf{Batch 4}} & \multicolumn{2}{c|}{\textbf{Batch 1}} & \multicolumn{2}{c}{\textbf{Batch 4}} \\
& E & \cellcolor{lightblue}P & E & \cellcolor{lightblue}P & E & \cellcolor{lightblue}P & E & \cellcolor{lightblue}P & E & \cellcolor{lightblue}P & E & \cellcolor{lightblue}P \\
\midrule
Non-Spec-Batch & -- & -- & 53.8 & \cellcolor{lightblue}98.2 & -- & -- & 92.9 & \cellcolor{lightblue}96.5 & -- & -- & 93.3 & \cellcolor{lightblue}97.2 \\
Spec-1         & 97.1 & \cellcolor{lightblue}98.4 & -- & -- & 94.6 & \cellcolor{lightblue}97.2 & -- & -- & 96.0 & \cellcolor{lightblue}98.0 & -- & -- \\
\oursb         & 97.3 & \cellcolor{lightblue}98.6 & 92.1 & \cellcolor{lightblue}98.6 & 94.6 & \cellcolor{lightblue}96.9 & 92.3 & \cellcolor{lightblue}95.7 & 96.7 & \cellcolor{lightblue}98.1 & 96.5 & \cellcolor{lightblue}98.3 \\
\oursx         & 97.3 & \cellcolor{lightblue}98.6 & 90.8 & \cellcolor{lightblue}97.6 & 94.6 & \cellcolor{lightblue}96.9 & 95.0 & \cellcolor{lightblue}97.1 & 96.7 & \cellcolor{lightblue}98.1 & 95.2 & \cellcolor{lightblue}97.7 \\
\cellcolor{lightred}DSD & \cellcolor{lightred}0.0 & \cellcolor{lightblue}8.1 & \cellcolor{lightred}0.0 & \cellcolor{lightblue}2.2 & \cellcolor{lightred}0.2 & \cellcolor{lightblue}2.2 & \cellcolor{lightred}0.0 & \cellcolor{lightblue}0.6 & \cellcolor{lightred}0.0 & \cellcolor{lightblue}1.0 & \cellcolor{lightred}0.0 & \cellcolor{lightblue}0.8 \\
\cellcolor{lightred}BSP & \cellcolor{lightred}1.9 & \cellcolor{lightblue}39.7 & \cellcolor{lightred}0.2 & \cellcolor{lightblue}31.3 & \cellcolor{lightred}3.5 & \cellcolor{lightblue}19.9 & \cellcolor{lightred}2.1 & \cellcolor{lightblue}12.6 & \cellcolor{lightred}1.0 & \cellcolor{lightblue}15.3 & \cellcolor{lightred}0.6 & \cellcolor{lightblue}8.1 \\
\bottomrule
\end{tabular}
\caption{Output equivalence of speculative-decoding implementations, measured by exact (E) and partial (P) match against batch-size-1 non-speculative decoding. \oursb and \oursx do not achieve bit-exact identity; DSD and BSP exhibit near-zero exact match.}
\label{tab:batch_based_verification}
\end{table*}

\begin{figure*}
    \centering
    \includegraphics[width=\linewidth]
    {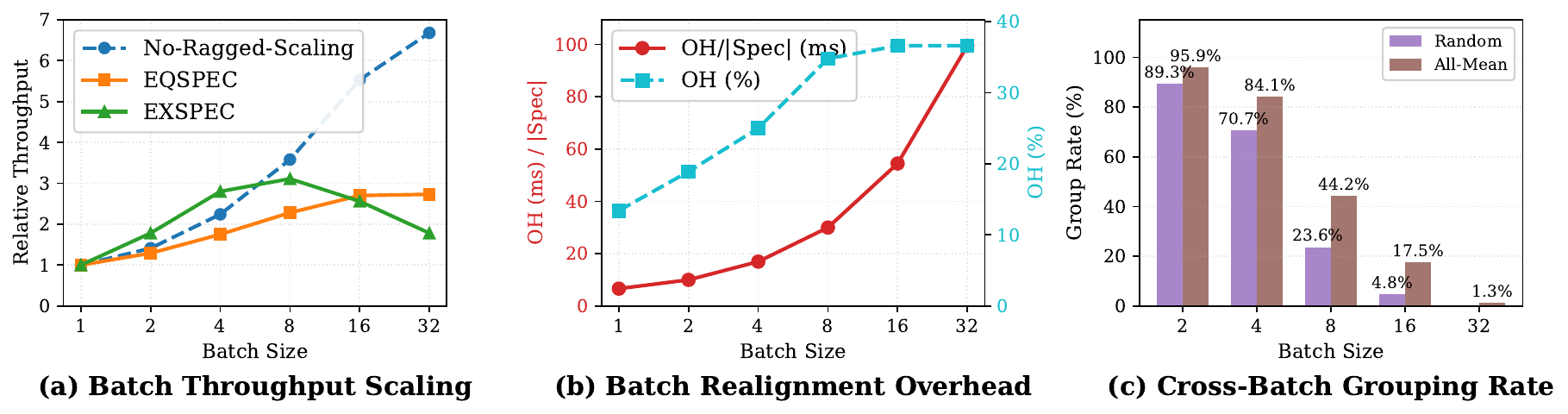}
\caption{
Decomposing batch speculative decoding performance. (a) Batch scaling efficiency: each method's throughput normalized to its own BS=1 baseline, isolating scaling behavior from absolute performance. The No-Ragged-Scaling line shows measured autoregressive decoding throughput (not a theoretical bound), serving as a reference for how standard batching scales without ragged tensor overhead.  (b) Alignment overhead grows superlinearly with batch size, consuming up to 38\% of inference time, validating that $c_{\text{overhead}}(B)$ dominates at scale. (c) Cross-batch grouping rates on Multi30k for random vs. uniform-length sequences, showing that length homogeneity transforms grouping effectiveness.}
    \label{fig:scaling_analysis}
    \vspace{-10pt}
\end{figure*} 


\subsection{Validating the Diagnostics (Stage 1--2)}
\label{sec:verification_exp}

The first two audit stages ask: do our diagnostics detect failures standard metrics miss, and do the signatures localize root causes? ROUGE~\citep{qian2024bass} is unsuitable as a correctness oracle: corrupted repetitive text can still score reasonably.

Table~\ref{tab:batch_based_verification} validates the diagnostics on two fronts. First, the metrics \emph{detect}: DSD and BSP show catastrophic degradation invisible to ROUGE. Second, they \emph{distinguish} root causes. DSD's near-zero scores on both metrics indicate immediate position-ID misalignment (failure from the first token). BSP shows higher partial match (up to 39.7\%) but low exact match: gradual decay as KV-cache drift misdirects attention, triggering repetition. These complementary signatures (immediate vs.\ gradual) map to the bug taxonomy (Table~\ref{tab:bug_taxonomy}): DSD suffers from batch-independent errors compounded by synchronization failures; BSP accumulates KV-cache drift from position-ID desynchronization.

\begin{table*}[t]
\centering
\small
\begin{tabular}{p{3.2cm}p{6.5cm}lcc}
\toprule
\textbf{Bug Category} & \textbf{Specific Issue} & \textbf{System} & \textbf{BS=1} & \textbf{BS$>$1} \\
\midrule
\multirow{2}{3.2cm}{Batch-independent implementation errors}
& Bonus token sampled from draft model instead of target model & DSD & \checkmark & \checkmark \\
& KV-cache contains rejected tokens; position IDs auto-calculated incorrectly & BSP & \checkmark & \checkmark \\
\midrule
\multirow{2}{3.2cm}{Ragged tensor synchronization (batch-specific)}
& Unaligned KV-cache across sequences with different accepted lengths & DSD, BSP & \ding{55} & \checkmark \\
& Position ID desynchronization across ragged sequences & BSP & \ding{55} & \checkmark \\
\bottomrule
\end{tabular}
\caption{Taxonomy of implementation errors in batch speculative decoding. Batch-independent errors corrupt outputs at any batch size and could be fixed with careful code review. Ragged tensor synchronization errors are batch-specific and represent the core challenge this paper addresses.}
\label{tab:bug_taxonomy}
\end{table*}
Enforcing the invariants removes the catastrophic failure signatures, but our methods achieve 90.8--97.3\% exact match against the batch-size-1 non-speculative oracle rather than bit-exact identity (Table~\ref{tab:batch_based_verification}). The remaining divergence is consistent with batch-dependent floating-point effects and argmax tie-breaking~\citep{he2025nondeterminism,gond2026llm42enablingdeterminismllm,liu2025speculative}; Table~\ref{tab:vllm_v1} shows that even non-speculative batched decoding can diverge from a batch-size-1 anchor.

\paragraph{vLLM~v1 numerical control.} Table~\ref{tab:vllm_v1} shows that non-speculative batched decoding also diverges from the batch-size-1 anchor when batch-invariant kernels are unavailable, separating numerical divergence from speculative-decoding synchronization. Because this vLLM configuration differs from our main setup, we treat it as a control rather than a matched performance comparison.

\begin{table}[t]
\centering
\footnotesize
\setlength{\tabcolsep}{2.5pt}
\begin{tabular}{llccc}
\toprule
\textbf{GPU} & \textbf{Batch-inv.} & \textbf{Non-spec} & \multicolumn{2}{c}{\textbf{Spec}} \\
 & & \textbf{BS$>$1} & \textbf{BS$=1$} & \textbf{BS$>$1} \\
\midrule
\multirow{2}{*}{H100} & on  & \textbf{80/80} & \textbf{80/80} & \textbf{80/80} \\
                      & off & 66/80 & 80/80 & 62/80 \\
\midrule
\multirow{2}{*}{A100} & on  & 62/80 & 58/80 & 56/80 \\
                      & off & 58/80 & 61/80 & 63/80 \\
\bottomrule
\end{tabular}
\caption{vLLM~v1 numerical control (Qwen3-4B/0.6B, $K{=}3$, 80 MT-Bench prompts, greedy): exact matches against the non-speculative BS=1 output.}
\label{tab:vllm_v1}
\end{table}


From a reproducibility perspective, these results underscore that published throughput numbers for batch speculative decoding should not be taken at face value without accompanying correctness verification. BSP and DSD reported competitive throughput in their respective publications, yet our audit reveals near-zero output equivalence: the measured tokens/s, while real, correspond to corrupted outputs rather than valid generation. We provide additional analysis of production systems (vLLM, SGLang) in Appendix~\ref{app:vllm_sglang}.

\subsection{Validating the Cost Analysis (Stage 4)}
\label{sec:overhead-and-scaling}

Five studies validate the cost predictions from the formalization stage: batch scaling efficiency, alignment overhead growth, sequence diversity as bottleneck, mechanistic profiling, and latency--throughput tradeoff.

\paragraph{Batch scaling efficiency.}
Figure~\ref{fig:scaling_analysis}(a) normalizes each method's throughput to its own BS=1 baseline, isolating batch-level scaling from absolute performance. The \emph{No-Ragged-Scaling} reference line is non-speculative decoding normalized the same way. \oursx initially exceeds this reference because speculation converts memory-bound generation into compute-bound verification that benefits more from GPU parallelism; at larger batch sizes, alignment overhead dominates and the advantage inverts. The key finding: valid batch speculative decoding need not sacrifice scaling efficiency up to BS=8, whereas prior methods either produce corrupted outputs or fail to scale.

\paragraph{Alignment overhead growth.} 

Figure~\ref{fig:scaling_analysis}(b) quantifies realignment costs via two metrics: percentage of total time spent on alignment (OH\%) and per-round alignment time (OH/\(|\text{Spec}|\)). Overhead rises from \(\sim13\%\) at BS=1 to nearly 40\% at BS=32, with per-round costs increasing even more. This matches our prediction that $c_{\text{pad}}(B)+c_{\text{kv}}(B)$ grows superlinearly with $B$. This is a structural cost of the contiguous layout: the very operations required to maintain the invariants (unpad--append--repad and KV-cache realignment) become increasingly expensive as sequence lengths diverge. The superlinear growth is not an implementation inefficiency but a property of maintaining synchronization under contiguous layouts (Theorem~\ref{thm:superlinear}), which helps explain why prior methods either break equivalence (skipping the expensive synchronization) or fail to scale (paying the full cost).

\paragraph{Grouping rate \texorpdfstring{$\times$}{x} sequence-length distribution.}

Figure~\ref{fig:scaling_analysis}(c) probes whether cross-batch scheduling limits are algorithmic or circumstantial via Multi30k. Comparing random sampling to an All-Mean subset with identical lengths isolates the bottleneck: sequence diversity, not the method. Random sampling shows grouping rates collapsing with batch size, whereas the All-Mean configuration maintains high grouping effectiveness even at moderate scales, substantially reducing $c_{\text{overhead}}(B)$. This contrast suggests that length-based bucketing or dynamic sorting can push real workloads toward the ideal. Production traces are length-heterogeneous~\citep{bai2026alignedserve}, and grouping similar-length requests is an established and effective serving technique~\citep{zheng2023response, bai2026alignedserve}; \oursx applies this principle under the synchronization invariants.

\begin{table}[t]
\centering
\footnotesize
\setlength{\tabcolsep}{3pt}
\begin{tabular}{l|rrr|rrr}
\toprule
Method & TPS & $|\text{Spec}|$& ms/V & V\% & D\% & OH\% \\
\midrule
\oursb & 95.6 & 1469 & 24.9 & 44.9 & 27.4 & 27.7\\
\oursx & 156.4 & 952 & 29.1 & 55.9 & 29.5 & 14.6\\
\bottomrule
\end{tabular}
\caption{Overhead anatomy of batch speculation methods. V\%=verification, D\%=draft, OH\%=overhead. Despite slower per-verification time, \oursx achieves higher throughput by reducing verification calls and alignment overhead.}
\vspace{-12pt}
\label{tab:anatomy}
\end{table}

\paragraph{Overhead Anatomy.} 

Table~\ref{tab:anatomy} shows \oursx achieves 64\% higher throughput by cutting verification calls by one-third and halving alignment overhead through same-length grouping. The trade-off is memory locality: dynamic batching scatters KV-cache entries, making each verification slower, but fewer total operations wins.

\paragraph{Latency--throughput tradeoff.} Simulating online serving with dynamic arrivals from shuffled multi-turn SpecBench conversations, Table~\ref{tab:latency} reveals two operating points. \oursx's grouping raises throughput but delays early requests until matching partners arrive, inflating tail latency, worst at small batch sizes where throughput is low; \oursb realigns every round for stable, predictable tail latency. The tradeoff is deployment-driven: \oursx for offline throughput, \oursb for latency-sensitive serving.

\begin{table}[t]
\centering
\small
\begin{tabular}{llrrr}
\toprule
BS & Method & Tok/s & P90 & P99 \\
\midrule
1 & \oursb & 15.20 & 12.96 & 16.53 \\
  & \oursx & 15.78 & 8.32  & 9.42  \\
\midrule
2 & \oursb & 26.70 & 14.78 & 19.06 \\
  & \oursx & 30.54 & 114.89 & 134.03 \\
\midrule
4 & \oursb & 46.11 & 16.23 & 19.13 \\
  & \oursx & 52.44 & 53.21 & 70.35 \\
\midrule
8 & \oursb & 76.03 & 19.23 & 22.50 \\
  & \oursx & 77.54 & 19.13 & 33.93 \\
\bottomrule
\end{tabular}
\caption{Latency--throughput tradeoff under simulated online serving with heterogeneous request lengths. Throughput in tokens/s; P90/P99 are the 90th/99th-percentile request-completion latency (s). \oursx maximizes throughput while \oursb keeps tail latency stable; setup and full breakdown in Appendix~\ref{app:latency}.}
\label{tab:latency}
\end{table}

\section{Related Work}
\label{sec:related-work}

\paragraph{Speculative Decoding and Batch Extension.}

Speculative decoding verifies draft tokens in parallel, either sequentially~\citep{xia:2022specdec,Santilli:2023paralleldecoding,Yang:2023PPD,Hooper:2023speed,Zhang:2023draftverify,Fu:2023lookahead} or via tree structures~\citep{Miao:2023specinfer,Spector:2023stagedspec,Sun:2023spectr,He:2023REST,medusa,li2024eagle,li2024eagle2,li2025eagle3scalinginferenceacceleration,tang2025efficient}. Exact variants preserve the target distribution; approximate schemes~\citep{Kim:2023bild,zhong2025speeding} trade exactness for speed. We assume lossless verification so that any deviation signals a bug. Recent work parallelizes multiple drafts per request~\citep{stewart2024n,lee2024inbatch} but still verifies independently.
Extending speculation to a batch of requests introduces the ragged tensor problem: when sequences accept different numbers of draft tokens, the resulting variable-length tensors break GPU-friendly rectangular operations~\citep{qian2024bass}. Existing approaches face fundamental limitations detailed in Section~\ref{sec:design-space}. Position ID/masking approaches like BSP~\citep{su2023synergy} suffer from position ID inconsistencies across iterations. Dynamic padding approaches reveal critical implementation errors: DSD~\citep{yan2025decoding} samples bonus tokens from the draft model's distribution rather than the target model's, violating output equivalence. This error compounds with improper KV-cache handling, producing corrupted outputs: BSP generates repetitive tokens, DSD produces \texttt{<unk>} symbols. BASS~\citep{qian2024bass} sidesteps these issues through custom CUDA kernels but sacrifices portability.
Production systems (vLLM, SGLang) avoid the problem through a different memory layout, continuous batching with variable-length and paged attention, rather than the synchronization we study (Appendix~\ref{app:vllm_sglang}). To our knowledge, no prior work formalizes the synchronization invariants that correctness requires; failures have been observed but not systematically analyzed.

\paragraph{Correctness Verification and Reproducibility in LLM Inference.}
Verifying the correctness of inference optimizations remains underexplored. \citet{liu2025speculative} report large gaps between theoretical and measured speculative-decoding speedup on vLLM and note output-length variation across batch sizes, attributing it to numerical non-determinism without further diagnosis; \citet{he2025nondeterminism} characterize floating-point non-determinism without addressing algorithmic correctness. Our diagnostic metrics separate this numerical noise from the synchronization failures we identify, drawing on fault-localization ideas from software engineering (spectrum-based localization~\citep{jones2005tarantula,abreu2007accuracy} and delta debugging~\citep{zeller2002simplifying}) that localize root causes from observable failure patterns.

\section{Conclusion}

Throughput without correctness verification is unreliable. Our forensic audit of batch speculative decoding followed four stages: \emph{observe} corrupted outputs in published implementations invisible to ROUGE; \emph{diagnose} via exact/partial match, distinguishing immediate position-ID failures from gradual KV-cache drift; \emph{formalize} the synchronization invariants whose violation produces each signature, proving superlinear realignment cost under contiguous layouts; \emph{validate} that enforcing the invariants removes catastrophic synchronization failures without bit-exact identity (90.8--97.3\% exact match), while \oursx cuts the cost via same-length scheduling ($3\times$ throughput at batch size 8). For practitioners: verify I1/I2 under greedy decoding; use exact match to detect corruption and partial match to localize it (high partial / low exact $\Rightarrow$ cache drift; both low $\Rightarrow$ position-ID errors). We hope the audit approach transfers to other tensor-batched settings, though validated here only for batch speculative decoding.


\section*{Limitations}
Our case study is limited to the contiguous KV-cache setting; we do not integrate \oursb or \oursx into continuous-batching systems such as vLLM or SGLang, whose variable-length and paged-attention layouts avoid the ragged-tensor problem by construction (Section~\ref{sec:design-space}). 
Within our case study, the advantage is confined to moderate batch sizes, since realignment overhead grows superlinearly and \oursx's gains depend on sequence-length homogeneity that heterogeneous workloads erode, while absolute speedup is bounded by draft-model quality. \oursx further trades tail latency for throughput: delaying requests to form same-length groups suits offline, throughput-oriented batch processing, whereas \oursb realigns each round and offers stable per-request latency for latency-sensitive online serving. We do not achieve bit-exact identity with batch-size-1 non-speculative decoding (90.8--97.3\% exact match); residual mismatches are consistent with batch-dependent floating-point effects.

\section*{Acknowledgments}
This work is supported by an eBay's Research and University Partnership for Technology (eRUPT) grant.
\bibliography{custom}

@article{bai2026alignedserve,
  title={AlignedServe: Orchestrating Prefix-aware Batching to Build a High-throughput and Computing-efficient LLM Serving System},
  author={Bai, Fengyao and Zhang, Hongbin and Chen, Zhitao and Du, Jiangsu and Chen, Zhiguang and Lu, Yutong},
  journal={Proceedings of the ACM on Management of Data},
  volume={4},
  number={3 (SIGMOD},
  pages={1--25},
  year={2026},
  publisher={ACM New York, NY, USA}
}

@article{zheng2023response,
  title={Response length perception and sequence scheduling: An llm-empowered llm inference pipeline},
  author={Zheng, Zangwei and Ren, Xiaozhe and Xue, Fuzhao and Luo, Yang and Jiang, Xin and You, Yang},
  journal={Advances in Neural Information Processing Systems},
  volume={36},
  pages={65517--65530},
  year={2023}
}

@article{liu2025speculative,
  title={Speculative Decoding: Performance or Illusion?},
  author={Liu, Xiaoxuan and Yu, Jiaxiang and Park, Jongseok and Stoica, Ion and Cheung, Alvin},
  journal={arXiv preprint arXiv:2601.11580},
  year={2026}
}

@misc{gond2026llm42enablingdeterminismllm,
      title={LLM-42: Enabling Determinism in LLM Inference with Verified Speculation}, 
      author={Raja Gond and Aditya K Kamath and Arkaprava Basu and Ramachandran Ramjee and Ashish Panwar},
      year={2026},
      eprint={2601.17768},
      archivePrefix={arXiv},
      primaryClass={cs.LG},
      url={https://arxiv.org/abs/2601.17768}, 
}

@article{su2024roformer,
  title={Roformer: Enhanced transformer with rotary position embedding},
  author={Su, Jianlin and Ahmed, Murtadha and Lu, Yu and Pan, Shengfeng and Bo, Wen and Liu, Yunfeng},
  journal={Neurocomputing},
  volume={568},
  pages={127063},
  year={2024},
  publisher={Elsevier}
}

@misc{lmsys_sglang_deterministic_2025,
  title        = {Enabling Deterministic Inference for SGLang},
  author       = {{SGLang Team}},
  year         = {2025},
  month        = sep,
  howpublished = {\url{https://lmsys.org/blog/2025-09-22-sglang-deterministic/}},
  note         = {LMSYS Org Blog, published Sep 22, 2025},
  urldate      = {2025-09-24}
}

@misc{rollback,
  author       = {kamilakesbi},
  title        = {Enable speculative decoding with batch size > 1 (GitHub issue \#32165)},
  howpublished = {\url{https://github.com/huggingface/transformers/issues/32165}},
  year         = {2024},
  note         = {Hugging Face \textit{transformers}. Opened 2024-07-23. Accessed 2025-09-24}
}

@InProceedings{multi30k,
  author = 	"Elliott, Desmond
		and Frank, Stella
		and Sima'an, Khalil
		and Specia, Lucia",
  title = 	"Multi30K: Multilingual English-German Image Descriptions",
  booktitle = 	"Proceedings of the 5th Workshop on Vision and Language",
  year = 	"2016",
  publisher = 	"Association for Computational Linguistics",
  pages = 	"70--74",
  location = 	"Berlin, Germany",
  doi = 	"10.18653/v1/W16-3210",
  url = 	"http://www.aclweb.org/anthology/W16-3210"
}

@inproceedings{wolf-etal-2020-transformers,
    title = "Transformers: State-of-the-Art Natural Language Processing",
    author = "Thomas Wolf and Lysandre Debut and Victor Sanh and Julien Chaumond and Clement Delangue and Anthony Moi and Pierric Cistac and Tim Rault and Rémi Louf and Morgan Funtowicz and Joe Davison and Sam Shleifer and Patrick von Platen and Clara Ma and Yacine Jernite and Julien Plu and Canwen Xu and Teven Le Scao and Sylvain Gugger and Mariama Drame and Quentin Lhoest and Alexander M. Rush",
    booktitle = "Proceedings of the 2020 Conference on Empirical Methods in Natural Language Processing: System Demonstrations",
    month = oct,
    year = "2020",
    address = "Online",
    publisher = "Association for Computational Linguistics",
    url = "https://www.aclweb.org/anthology/2020.emnlp-demos.6",
    pages = "38--45"
}

@article{zhong2025speeding,
  title={Speeding up Speculative Decoding via Sequential Approximate Verification},
  author={Zhong, Meiyu and Teku, Noel and Tandon, Ravi},
  journal={arXiv preprint arXiv:2502.04557},
  year={2025}
}

@article{he2025nondeterminism,
  author = {Horace He and Thinking Machines Lab},
  title = {Defeating Nondeterminism in LLM Inference},
  journal = {Thinking Machines Lab: Connectionism},
  year = {2025},
  note = {https://thinkingmachines.ai/blog/defeating-nondeterminism-in-llm-inference/},
  doi = {10.64434/tml.20250910}
}

@misc{liu2025turbospecclosedloopspeculationcontrol,
      title={TurboSpec: Closed-loop Speculation Control System for Optimizing LLM Serving Goodput}, 
      author={Xiaoxuan Liu and Jongseok Park and Langxiang Hu and Woosuk Kwon and Zhuohan Li and Chen Zhang and Kuntai Du and Xiangxi Mo and Kaichao You and Alvin Cheung and Zhijie Deng and Ion Stoica and Hao Zhang},
      year={2025},
      eprint={2406.14066},
      archivePrefix={arXiv},
      primaryClass={cs.AI},
      url={https://arxiv.org/abs/2406.14066}, 
}

@inproceedings{wu2024tetris,
  title={TETRIS: Optimal Draft Token Selection for Batch Speculative Decoding},
  author={Wu, Zhaoxuan and Zhou, Zijian and Verma, Arun and Prakash, Alok and Rus, Daniela and Low, Bryan Kian Hsiang},
  booktitle={Proceedings of the 63rd Annual Meeting of the Association for Computational Linguistics (ACL)},
  year={2025}
}

@inproceedings{kwon2023efficient,
  title={Efficient Memory Management for Large Language Model Serving with PagedAttention},
  author={Woosuk Kwon and Zhuohan Li and Siyuan Zhuang and Ying Sheng and Lianmin Zheng and Cody Hao Yu and Joseph E. Gonzalez and Hao Zhang and Ion Stoica},
  booktitle={Proceedings of the ACM SIGOPS 29th Symposium on Operating Systems Principles},
  year={2023}
}

@article{zheng2024sglang,
  title={Sglang: Efficient execution of structured language model programs},
  author={Zheng, Lianmin and Yin, Liangsheng and Xie, Zhiqiang and Sun, Chuyue Livia and Huang, Jeff and Yu, Cody Hao and Cao, Shiyi and Kozyrakis, Christos and Stoica, Ion and Gonzalez, Joseph E and others},
  journal={Advances in neural information processing systems},
  volume={37},
  pages={62557--62583},
  year={2024}
}

@misc{glm2024chatglm,
      title={ChatGLM: A Family of Large Language Models from GLM-130B to GLM-4 All Tools},
      author={Team GLM and Aohan Zeng and Bin Xu and Bowen Wang and Chenhui Zhang and Da Yin and Diego Rojas and Guanyu Feng and Hanlin Zhao and Hanyu Lai and Hao Yu and Hongning Wang and Jiadai Sun and Jiajie Zhang and Jiale Cheng and Jiayi Gui and Jie Tang and Jing Zhang and Juanzi Li and Lei Zhao and Lindong Wu and Lucen Zhong and Mingdao Liu and Minlie Huang and Peng Zhang and Qinkai Zheng and Rui Lu and Shuaiqi Duan and Shudan Zhang and Shulin Cao and Shuxun Yang and Weng Lam Tam and Wenyi Zhao and Xiao Liu and Xiao Xia and Xiaohan Zhang and Xiaotao Gu and Xin Lv and Xinghan Liu and Xinyi Liu and Xinyue Yang and Xixuan Song and Xunkai Zhang and Yifan An and Yifan Xu and Yilin Niu and Yuantao Yang and Yueyan Li and Yushi Bai and Yuxiao Dong and Zehan Qi and Zhaoyu Wang and Zhen Yang and Zhengxiao Du and Zhenyu Hou and Zihan Wang},
      year={2024},
      eprint={2406.12793},
      archivePrefix={arXiv},
      primaryClass={id='cs.CL' full_name='Computation and Language' is_active=True alt_name='cmp-lg' in_archive='cs' is_general=False description='Covers natural language processing. Roughly includes material in ACM Subject Class I.2.7. Note that work on artificial languages (programming languages, logics, formal systems) that does not explicitly address natural-language issues broadly construed (natural-language processing, computational linguistics, speech, text retrieval, etc.) is not appropriate for this area.'}
}

@article{qwen3,
    title={Qwen3 Technical Report}, 
    author={An Yang and Anfeng Li and Baosong Yang and Beichen Zhang and Binyuan Hui and Bo Zheng and Bowen Yu and Chang Gao and Chengen Huang and Chenxu Lv and Chujie Zheng and Dayiheng Liu and Fan Zhou and Fei Huang and Feng Hu and Hao Ge and Haoran Wei and Huan Lin and Jialong Tang and Jian Yang and Jianhong Tu and Jianwei Zhang and Jianxin Yang and Jiaxi Yang and Jing Zhou and Jingren Zhou and Junyang Lin and Kai Dang and Keqin Bao and Kexin Yang and Le Yu and Lianghao Deng and Mei Li and Mingfeng Xue and Mingze Li and Pei Zhang and Peng Wang and Qin Zhu and Rui Men and Ruize Gao and Shixuan Liu and Shuang Luo and Tianhao Li and Tianyi Tang and Wenbiao Yin and Xingzhang Ren and Xinyu Wang and Xinyu Zhang and Xuancheng Ren and Yang Fan and Yang Su and Yichang Zhang and Yinger Zhang and Yu Wan and Yuqiong Liu and Zekun Wang and Zeyu Cui and Zhenru Zhang and Zhipeng Zhou and Zihan Qiu},
    journal = {arXiv preprint arXiv:2505.09388},
    year={2025}
}

@article{zheng2023judging,
  title={Judging LLM-as-a-judge with MT-Bench and Chatbot Arena},
  author={Lianmin Zheng and Wei-Lin Chiang and Ying Sheng and Siyuan Zhuang and Zhanghao Wu and Yonghao Zhuang and Zi Lin and Zhuohan Li and Dacheng Li and Eric. P Xing and Hao Zhang and Joseph E. Gonzalez and Ion Stoica},
  journal={Advances in Neural Information Processing Systems},
  volume={36},
  pages={46595--46623},
  year={2023}
}

@inproceedings{xia-etal-2024-unlocking,
    title = "Unlocking Efficiency in Large Language Model Inference: A Comprehensive Survey of Speculative Decoding",
    author = "Xia, Heming and Yang, Zhe and Dong, Qingxiu and Wang, Peiyi and Li, Yongqi  and Ge, Tao and Liu, Tianyu and Li, Wenjie and Sui, Zhifang",
    editor = "Ku, Lun-Wei and Martins, Andre and Srikumar, Vivek",
    booktitle = "Findings of the Association for Computational Linguistics ACL 2024",
    month = aug,
    year = "2024",
    address = "Bangkok, Thailand and virtual meeting",
    publisher = "Association for Computational Linguistics",
    url = "https://aclanthology.org/2024.findings-acl.456",
    doi = "10.18653/v1/2024.findings-acl.456",
    pages = "7655--7671",
}

@inproceedings{li2024eagle,
  title={EAGLE: Speculative Sampling Requires Rethinking Feature Uncertainty},
  author={Li, Yuhui and Wei, Fangyun and Zhang, Chao and Zhang, Hongyang},
  booktitle={International Conference on Machine Learning},
  pages={28935--28948},
  year={2024},
  organization={PMLR}
}

@inproceedings{li2024eagle2,
  title={EAGLE-2: Faster Inference of Language Models with Dynamic Draft Trees},
  author={Li, Yuhui and Wei, Fangyun and Zhang, Chao and Zhang, Hongyang},
  booktitle={Proceedings of the 2024 Conference on Empirical Methods in Natural Language Processing},
  pages={7421--7432},
  year={2024}
}

@article{li2025eagle3scalinginferenceacceleration,
  title={{EAGLE-3}: Scaling up Inference Acceleration of Large Language Models via Training-Time Test},
  author={Yuhui Li and Fangyun Wei and Chao Zhang and Hongyang Zhang},
  journal={Advances in Neural Information Processing Systems},
  volume={38},
  pages={136737--136756},
  year={2026}
}

@article{tang2025efficient,
  title={Efficient Speculative Decoding for Llama at Scale: Challenges and Solutions},
  author={Tang, Bangsheng and Fu, Carl Chengyan and Kou, Fei and Sizov, Grigory and Zhang, Haoci and Park, Jason and Liu, Jiawen and You, Jie and Yang, Qirui and Mehta, Sachin and others},
  journal={arXiv preprint arXiv:2508.08192},
  year={2025}
}

@inproceedings{leviathan2023fast,
  title={Fast inference from transformers via speculative decoding},
  author={Leviathan, Yaniv and Kalman, Matan and Matias, Yossi},
  booktitle={International Conference on Machine Learning},
  pages={19274--19286},
  year={2023},
  organization={PMLR}
}

@article{chen2023accelerating,
  title={Accelerating large language model decoding with speculative sampling},
  author={Chen, Charlie and Borgeaud, Sebastian and Irving, Geoffrey and Lespiau, Jean-Baptiste and Sifre, Laurent and Jumper, John},
  journal={arXiv preprint arXiv:2302.01318},
  year={2023}
}

@inproceedings{qian2024bass,
  title={BASS: Batched Attention-optimized Speculative Sampling},
  author={Qian, Haifeng and Gonugondla, Sujan Kumar and Ha, Sungsoo and Shang, Mingyue and Gouda, Sanjay Krishna and Nallapati, Ramesh and Sengupta, Sudipta and Ma, Xiaofei and Deoras, Anoop},
  booktitle={Findings of the Association for Computational Linguistics ACL 2024},
  pages={8214--8224},
  year={2024}
}

@inproceedings{yan2025decoding,
  title={Decoding Speculative Decoding},
  author={Yan, Minghao and Agarwal, Saurabh and Venkataraman, Shivaram},
  booktitle={Proceedings of the 2025 Conference of the Nations of the Americas Chapter of the Association for Computational Linguistics: Human Language Technologies (Volume 1: Long Papers)},
  pages={6460--6473},
  year={2025}
}

@article{su2023synergy,
  title={The synergy of speculative decoding and batching in serving large language models},
  author={Su, Qidong and Giannoula, Christina and Pekhimenko, Gennady},
  journal={arXiv preprint arXiv:2310.18813},
  year={2023}
}

@article{stewart2024n,
  title={The n-grammys: Accelerating autoregressive inference with learning-free batched speculation},
  author={Stewart, Lawrence and Trager, Matthew and Gonugondla, Sujan Kumar and Soatto, Stefano},
  journal={arXiv preprint arXiv:2411.03786},
  year={2024}
}

@misc{lee2024inbatch,
    title={In-batch Ensemble Drafting: Toward Fast and Robust Speculative Decoding for Multimodal Language Models},
    author={Minjae Lee and Wonjun Kang and Minghao Yan and Christian Classen and Hyung Il Koo and Kangwook Lee},
    year={2024},
    url={https://openreview.net/forum?id=8o7131Lm83}
}

@inproceedings{xia:2022specdec,
  author       = {Heming Xia and
                  Tao Ge and
                  Peiyi Wang and
                  Si{-}Qing Chen and
                  Furu Wei and
                  Zhifang Sui},
  editor       = {Houda Bouamor and
                  Juan Pino and
                  Kalika Bali},
  title        = {Speculative Decoding: Exploiting Speculative Execution for Accelerating
                  Seq2seq Generation},
  booktitle    = {Findings of the Association for Computational Linguistics: {EMNLP}
                  2023, Singapore, December 6-10, 2023},
  pages        = {3909--3925},
  publisher    = {Association for Computational Linguistics},
  year         = {2023},
  url          = {https://doi.org/10.18653/v1/2023.findings-emnlp.257},
  doi          = {10.18653/V1/2023.FINDINGS-EMNLP.257},
  bibsource    = {dblp computer science bibliography, https://dblp.org}
}

@inproceedings{Santilli:2023paralleldecoding,
  author       = {Andrea Santilli and
                  Silvio Severino and
                  Emilian Postolache and
                  Valentino Maiorca and
                  Michele Mancusi and
                  Riccardo Marin and
                  Emanuele Rodol{\`{a}}},
  editor       = {Anna Rogers and
                  Jordan L. Boyd{-}Graber and
                  Naoaki Okazaki},
  title        = {Accelerating Transformer Inference for Translation via Parallel Decoding},
  booktitle    = {Proceedings of the 61st Annual Meeting of the Association for Computational
                  Linguistics (Volume 1: Long Papers), {ACL} 2023, Toronto, Canada,
                  July 9-14, 2023},
  pages        = {12336--12355},
  publisher    = {Association for Computational Linguistics},
  year         = {2023},
  url          = {https://doi.org/10.18653/v1/2023.acl-long.689},
  doi          = {10.18653/v1/2023.acl-long.689},
  bibsource    = {dblp computer science bibliography, https://dblp.org}
}

@article{Yang:2023PPD,
  author       = {Seongjun Yang and
                  Gibbeum Lee and
                  Jaewoong Cho and
                  Dimitris S. Papailiopoulos and
                  Kangwook Lee},
  title        = {Predictive Pipelined Decoding: {A} Compute-Latency Trade-off for Exact
                  {LLM} Decoding},
  journal      = {CoRR},
  volume       = {abs/2307.05908},
  year         = {2023},
  url          = {https://doi.org/10.48550/arXiv.2307.05908},
  doi          = {10.48550/ARXIV.2307.05908},
  eprinttype    = {arXiv},
  eprint       = {2307.05908},
  bibsource    = {dblp computer science bibliography, https://dblp.org}
}

@article{Hooper:2023speed,
  author       = {Coleman Hooper and
                  Sehoon Kim and
                  Hiva Mohammadzadeh and
                  Hasan Genc and
                  Kurt Keutzer and
                  Amir Gholami and
                  Yakun Sophia Shao},
  title        = {{SPEED:} Speculative Pipelined Execution for Efficient Decoding},
  journal      = {CoRR},
  volume       = {abs/2310.12072},
  year         = {2023},
  url          = {https://doi.org/10.48550/arXiv.2310.12072},
  doi          = {10.48550/ARXIV.2310.12072},
  eprinttype    = {arXiv},
  eprint       = {2310.12072},
  bibsource    = {dblp computer science bibliography, https://dblp.org}
}

@inproceedings{Zhang:2023draftverify,
  title        = {Draft {\&} Verify: Lossless Large Language Model Acceleration via Self-Speculative Decoding},
  author       = {Jun Zhang and Jue Wang and Huan Li and Lidan Shou and Ke Chen and Gang Chen and Sharad Mehrotra},
  booktitle    = {Proceedings of the 62nd Annual Meeting of the Association for Computational Linguistics (Volume 1: Long Papers)},
  pages        = {11263--11282},
  year         = {2024}
}

@misc{Fu:2023lookahead,
      title={Break the Sequential Dependency of LLM Inference Using Lookahead Decoding}, 
      author={Yichao Fu and Peter Bailis and Ion Stoica and Hao Zhang},
      year={2024},
      eprint={2402.02057},
      archivePrefix={arXiv},
      primaryClass={cs.LG}
}

@inproceedings{Kim:2023bild,
  author       = {Sehoon Kim and
                  Karttikeya Mangalam and
                  Suhong Moon and
                  Jitendra Malik and
                  Michael W. Mahoney and
                  Amir Gholami and
                  Kurt Keutzer},
  editor       = {Alice Oh and
                  Tristan Naumann and
                  Amir Globerson and
                  Kate Saenko and
                  Moritz Hardt and
                  Sergey Levine},
  title        = {Speculative Decoding with Big Little Decoder},
  booktitle    = {Advances in Neural Information Processing Systems 36: Annual Conference
                  on Neural Information Processing Systems 2023, NeurIPS 2023, New Orleans,
                  LA, USA, December 10 - 16, 2023},
  year         = {2023},
  url          = {http://papers.nips.cc/paper\_files/paper/2023/hash/7b97adeafa1c51cf65263459ca9d0d7c-Abstract-Conference.html},
  bibsource    = {dblp computer science bibliography, https://dblp.org}
}

@inproceedings{Miao:2023specinfer,
    author = {Miao, Xupeng and Oliaro, Gabriele and Zhang, Zhihao and Cheng, Xinhao and Wang, Zeyu and Zhang, Zhengxin and Wong, Rae Ying Yee and Zhu, Alan and Yang, Lijie and Shi, Xiaoxiang and Shi, Chunan and Chen, Zhuoming and Arfeen, Daiyaan and Abhyankar, Reyna and Jia, Zhihao},
    title = {SpecInfer: Accelerating Large Language Model Serving with Tree-based Speculative Inference and Verification},
    year = {2024},
    isbn = {9798400703867},
    publisher = {Association for Computing Machinery},
    address = {New York, NY, USA},
    url = {https://doi.org/10.1145/3620666.3651335},
    doi = {10.1145/3620666.3651335},
    booktitle = {Proceedings of the 29th ACM International Conference on Architectural Support for Programming Languages and Operating Systems, Volume 3},
    pages = {932–949},
    numpages = {18},
    location = {<conf-loc>, <city>La Jolla</city>, <state>CA</state>, <country>USA</country>, </conf-loc>},
    series = {ASPLOS '24}
}

@article{Spector:2023stagedspec,
  author       = {Benjamin Spector and
                  Chris Re},
  title        = {Accelerating {LLM} Inference with Staged Speculative Decoding},
  journal      = {CoRR},
  volume       = {abs/2308.04623},
  year         = {2023},
  url          = {https://doi.org/10.48550/arXiv.2308.04623},
  doi          = {10.48550/arXiv.2308.04623},
  eprinttype    = {arXiv},
  eprint       = {2308.04623},
  bibsource    = {dblp computer science bibliography, https://dblp.org}
}

@inproceedings{Sun:2023spectr,
  author       = {Ziteng Sun and
                  Ananda Theertha Suresh and
                  Jae Hun Ro and
                  Ahmad Beirami and
                  Himanshu Jain and
                  Felix X. Yu},
  editor       = {Alice Oh and
                  Tristan Naumann and
                  Amir Globerson and
                  Kate Saenko and
                  Moritz Hardt and
                  Sergey Levine},
  title        = {SpecTr: Fast Speculative Decoding via Optimal Transport},
  booktitle    = {Advances in Neural Information Processing Systems 36: Annual Conference
                  on Neural Information Processing Systems 2023, NeurIPS 2023, New Orleans,
                  LA, USA, December 10 - 16, 2023},
  year         = {2023},
  url          = {http://papers.nips.cc/paper\_files/paper/2023/hash/6034a661584af6c28fd97a6f23e56c0a-Abstract-Conference.html},
  bibsource    = {dblp computer science bibliography, https://dblp.org}
}

@inproceedings{He:2023REST,
  title        = {{REST:} Retrieval-Based Speculative Decoding},
  author       = {Zhenyu He and Zexuan Zhong and Tianle Cai and Jason D. Lee and Di He},
  booktitle    = {Proceedings of the 2024 Conference of the North American Chapter of the Association for Computational Linguistics: Human Language Technologies (Volume 1: Long Papers)},
  pages        = {1582--1595},
  year         = {2024}
}

@article{medusa,
  author       = {Tianle Cai and
                  Yuhong Li and
                  Zhengyang Geng and
                  Hongwu Peng and
                  Jason D. Lee and
                  Deming Chen and
                  Tri Dao},
  title        = {Medusa: Simple {LLM} Inference Acceleration Framework with Multiple
                  Decoding Heads},
  journal      = {CoRR},
  volume       = {abs/2401.10774},
  year         = {2024},
  url          = {https://doi.org/10.48550/arXiv.2401.10774},
  doi          = {10.48550/ARXIV.2401.10774},
  eprinttype    = {arXiv},
  eprint       = {2401.10774},
  bibsource    = {dblp computer science bibliography, https://dblp.org}
}

@article{zeller2002simplifying,
  title={Simplifying and Isolating Failure-Inducing Input},
  author={Zeller, Andreas and Hildebrandt, Ralf},
  journal={IEEE Transactions on Software Engineering},
  volume={28},
  number={2},
  pages={183--200},
  year={2002},
  publisher={IEEE}
}

@inproceedings{jones2005tarantula,
  title={Empirical Evaluation of the Tarantula Automatic Fault-Localization Technique},
  author={Jones, James A. and Harrold, Mary Jean},
  booktitle={Proceedings of the 20th IEEE/ACM International Conference on Automated Software Engineering},
  pages={273--282},
  year={2005},
  organization={ACM}
}

@inproceedings{abreu2007accuracy,
  title={On the Accuracy of Spectrum-based Fault Localization},
  author={Abreu, Rui and Zoeteweij, Peter and Van Gemund, Arjan J.C.},
  booktitle={Testing: Academic and Industrial Conference Practice and Research Techniques - MUTATION},
  pages={89--98},
  year={2007},
  organization={IEEE}
}

\appendix

\subsection*{Appendix Contents}
\vspace{-0.5em}
\begin{itemize}[leftmargin=1.5em, itemsep=0pt, parsep=1pt, topsep=0pt]
    \item[{\scriptsize\ref{app:bug_taxonomy}}] Bug Taxonomy in Prior Implementations
    \item[{\scriptsize\ref{app:proof}}] Proof of Synchronization Invariants
    \item[{\scriptsize\ref{app:exspec}}] \oursx Cross-Batch Scheduling
    \item[{\scriptsize\ref{app:batchverify}}] BatchVerify Algorithm Details
    \item[{\scriptsize\ref{app:latency}}] Latency-Throughput Tradeoff in Online Serving
    \item[{\scriptsize\ref{app:vllm_sglang}}] Continuous Batching Systems
    \item[{\scriptsize\ref{app:discussion}}] Discussion
    \item[{\scriptsize\ref{app:reproducibility}}] Reproducibility Statement
    \item[{\scriptsize\ref{app:llm}}] Disclose on LLM Usage
\end{itemize}
\vspace{0.5em}

\section{Bug Taxonomy in Prior Implementations}
\label{app:bug_taxonomy}

We categorize the implementation errors found in prior batch speculative decoding systems by whether they manifest at batch size~1 (Table~\ref{tab:bug_taxonomy}).

\paragraph{Batch-independent errors.} The first category consists of relatively straightforward implementation mistakes: sampling from the wrong distribution or failing to exclude rejected tokens. While these corrupt outputs at any batch size, they could be fixed with careful code review. DSD incorrectly samples the bonus token from the draft model's distribution rather than the target model's, violating the fundamental output equivalence guarantee of speculative decoding. BSP's KV-cache retains rejected tokens and relies on framework auto-calculation for position IDs, which produces incorrect values when padding patterns change.

\paragraph{Batch-specific synchronization errors.} The second category represents the core contribution of this paper: errors that \emph{only} manifest when batch size exceeds one. These arise from the ragged tensor problem: when sequences accept different numbers of draft tokens, maintaining synchronized position IDs, attention masks, and KV-cache state across the batch becomes non-trivial. Our synchronization invariants (I1, I2) in Section~\ref{sec:method} formalize exactly what must be preserved, and our \oursb algorithm provides the first correct implementation.

This taxonomy clarifies that while prior implementations suffer from multiple bug types, our contribution specifically addresses the batch-specific synchronization challenge that has no analog in single-sequence speculative decoding.

\section{Proof of Synchronization Invariants}
\label{app:proof}

\begin{theorem}[\oursb: Preservation of Alignment and Cache Correspondence]
Given a batch of size $B$, let $p_i^{(t)}$ and $c_i^{(t)}$ denote the padding and content lengths of sequence $i$ at verification step $t$, and let realignment apply the offset $\delta_i$ of Eq.~(2) as a \emph{relocation}: a value-preserving move of cached entries, with rejected entries dropped. If at step $t$ (i) the alignment invariant $p_i^{(t)} + c_i^{(t)} = L^{(t)}$ holds for all $i \in [1,B]$, and (ii) the cache clause of I2 holds, so every content position stores the key/value the target model computed for its token, then the \oursb update rule guarantees both properties at step $t+1$. The position-ID clause of I2 is restored directly by recomputing position IDs from the attention mask after repadding.
\label{theorem}
\end{theorem}

\begin{proof}
We prove this by induction on the verification step $t$.

\textbf{Base Case ($t=0$):}
At initialization, all sequences are left-padded to the maximum prompt length $L^{(0)}$. Thus, $\forall i, p_i^{(0)} + c_i^{(0)} = L^{(0)}$. The invariant holds. At $t=0$ the cache holds only the prompt entries from the initial target forward pass, which are correct by construction, so (ii) holds as well.

\textbf{Inductive Step:}
Assume the invariant holds at step $t$. During verification, sequence $i$ accepts $a_i$ new tokens. The new content length becomes:
\begin{equation}
    c_i^{(t+1)} = c_i^{(t)} + a_i
\end{equation}
The new target batch length is determined by the longest updated sequence:
\begin{equation}
    L^{(t+1)} = \max_{j \in [1, B]} (c_j^{(t)} + a_j)
\end{equation}
The EQSPEC algorithm calculates the padding offset $\delta_i$ as defined in Eq. (2):
\begin{equation}
    \delta_i = (L^{(t+1)} - L^{(t)}) - a_i
\end{equation}
The new padding length is updated as $p_i^{(t+1)} = p_i^{(t)} + \delta_i$. We verify the invariant at $t+1$ by substituting these terms:
\begin{align*}
    p_i^{(t+1)} + c_i^{(t+1)} &= (p_i^{(t)} + \delta_i) + (c_i^{(t)} + a_i) \\
    &= p_i^{(t)} + (L^{(t+1)} - L^{(t)} - a_i) + c_i^{(t)} + a_i \\
    &= (p_i^{(t)} + c_i^{(t)}) - L^{(t)} + L^{(t+1)}
\end{align*}
By the inductive hypothesis, $p_i^{(t)} + c_i^{(t)} = L^{(t)}$. Substituting this yields:
\begin{align*}
    &= L^{(t)} - L^{(t)} + L^{(t+1)} \\
    &= L^{(t+1)}
\end{align*}
Thus, $\forall i, p_i^{(t+1)} + c_i^{(t+1)} = L^{(t+1)}$. The rectangular alignment is preserved.

\textbf{Cache clause of I2.} \oursb realignment is a relocation: it moves each retained entry to the offset induced by $\delta_i$ without recomputation, drops the entries of rejected draft tokens, and keeps the entries the target produced for accepted tokens during verification. Relocation leaves stored values unchanged, so each retained entry still equals the target-computed value from the round in which its token was processed; by the inductive hypothesis this is the value required at $t+1$. This round's accepted tokens take their entries directly from the verification pass, and rejected tokens contribute none. Hence every content position holds the correct key/value at $t+1$. (The bonus token carries no entry until the next forward pass; see Appendix~\ref{app:batchverify}.)
\end{proof}

\begin{corollary}[Output Equivalence under Exact Arithmetic]
If a batched run maintains I1 and I2 at every round and samples each bonus token from the target model, then under exact arithmetic and a deterministic policy each sequence produces the same tokens as standard per-sequence decoding.
\label{cor:equiv}
\end{corollary}

\begin{proof}
At each round the target reads, for every sequence, the reference positions and keys/values (I2) within a rectangular batch (I1). Under identical arithmetic, it therefore computes the same logits as standard decoding. A deterministic policy then selects the same accepted tokens and the same bonus token. Induction over rounds yields identical token sequences.
\end{proof}

Finite-precision batched kernels may alter argmax decisions; Section~\ref{sec:verification_exp} and Table~\ref{tab:vllm_v1} evaluate this numerical divergence separately.

\begin{corollary}[Zero Overhead for Single Sequence]
For batch size $B=1$, no padding adjustment is required.
\label{corollary_hf}
\end{corollary}

\begin{proof}
When $B=1$, the batch consists of a single sequence with content length $c_1^{(t)}$. We can set padding $p_1^{(t)} = 0$ without loss of generality, satisfying the alignment invariant naturally as $L^{(t)} = c_1^{(t)}$.
Given acceptance $a_1$, the new length is $L^{(t+1)} = c_1^{(t)} + a_1$.
Substituting into Eq. (2):
\begin{equation}
    \delta_1 = (L^{(t+1)} - L^{(t)}) - a_1 = (c_1^{(t)} + a_1 - c_1^{(t)}) - a_1 = 0
\end{equation}
Since $\delta_1 = 0$, no unpad-append-repad or KV-cache shifting is required.
\end{proof}



\begin{corollary}[\oursx: Zero Overhead for Uniform Acceptance]
If all sequences in a batch accept the same number of tokens $k \ge 1$, no padding realignment is required ($\delta_i = 0$ for all $i$), regardless of their initial padding or content lengths.
\label{corollary_xbatch}
\end{corollary}

\begin{proof}
Let any sequence $i$ have arbitrary padding $p_i^{(t)}$ and content $c_i^{(t)}$.
The batch length at step $t$ is defined as $L^{(t)} = \max_j (p_j^{(t)} + c_j^{(t)})$.
Recall that the invariant $p_j^{(t)} + c_j^{(t)} = L^{(t)}$ holds for all sequences.

If every sequence accepts an identical number of tokens $a_i = k$, then the new content length is $c_i^{(t+1)} = c_i^{(t)} + k$.
The new batch length becomes:
\begin{align*}
    L^{(t+1)} &= \max_j (p_j^{(t)} + c_j^{(t+1)}) \\
    &= \max_j (p_j^{(t)} + c_j^{(t)} + k) \\
    &= (\max_j (p_j^{(t)} + c_j^{(t)})) + k \\
    &= L^{(t)} + k
\end{align*}
The realignment offset $\delta_i$ is calculated as:
\begin{align*}
    \delta_i &= (L^{(t+1)} - L^{(t)}) - a_i \\
    &= ((L^{(t)} + k) - L^{(t)}) - k \\
    &= k - k \\
    &= 0
\end{align*}
Since $\delta_i = 0$, the existing padding structure is preserved exactly. No memory reallocation or shifting is required, even if the sequences started with different amounts of padding.
\end{proof}

\begin{proof}[Proof of Theorem~\ref{thm:superlinear}]
Write $C(B)=\sum_{i=1}^{B}(M_B-a_i)$ for the per-round realignment work. By linearity of expectation, $E[C(B)]=B\,(E[M_B]-\mu)$. Adding a sequence can only raise the running maximum, so $E[M_{B+1}]\ge E[M_B]$, with strict inequality whenever acceptances are non-degenerate; the per-sequence cost $E[M_B]-\mu$ is therefore increasing in $B$, and the total $B\,(E[M_B]-\mu)$ is superadditive (it grows strictly faster than linearly). Because acceptances are capped at $K$ and some sequence attains $K$ with positive probability, $P(M_B<K)=[1-P(a_i=K)]^B\to0$, so $E[M_B]\to K$ and the per-sequence cost saturates at $K-\mu$.
\end{proof}

\section{\oursx: Cross-Batch Scheduling Algorithm}
\label{app:exspec}
Algorithm~\ref{alg:cross_batch_speculative} gives the complete \oursx procedure summarized in Section~\ref{sec:algo_analysis}. It maintains a \emph{SequencePool} of active sequences and, over a sliding window of width $W > B$, forms batches of identical length that concatenate without realignment, falling back to the unpad--append--repad path of \oursb only when same-length grouping fails.
\begin{algorithm}[tb]
\footnotesize
\caption{\oursx: Cross-Batch Scheduling}
\label{alg:cross_batch_speculative}
\begin{algorithmic}[1]
\REQUIRE Draft and Target model $\mathcal{M}_d$, $\mathcal{M}_t$, Prompts $\mathcal{P}$, Window size $W$, Batch size $B$
\ENSURE Generated sequences $\mathcal{S}$
\STATE $\textit{Pool} \gets \texttt{InitSequencePool}(\mathcal{P})$
\STATE \COMMENT{Tokenize and optionally sort by length}
\STATE $\textit{Window} \gets \texttt{RefillWindow}(\textit{Pool}, W)$
\WHILE{$\textit{Pool.hasActive()}$}
    \STATE \textbf{Phase 1: Lazy Realignment}
    \STATE $ \mathcal{B}, \textit{mask}, \textit{KV} \gets \texttt{GetBatch}(\textit{Window}, B)$
    \STATE \COMMENT{Try same-length concatenation}
    \STATE \COMMENT{Fallback to Unpad-Repad Realignment}
    \STATE \textbf{Phase 2: Draft Generation}
    \STATE $D \gets \mathcal{M}_d.\texttt{Generate}(\mathcal{B}, \textit{mask}, K)$
    \STATE \textbf{Phase 3: Batch Verification}
    \STATE \COMMENT{See Alg. 3 in Appendix for BatchVerify}
    \STATE $A, B, \textit{KV} \gets \texttt{BatchVerify}(\mathcal{M}_t, \mathcal{B}, D, \textit{KV})$
    \STATE \textbf{Phase 4: Write-Back and Window Refill}
    \FOR{$i \in \mathcal{B}$}
        \STATE $\textit{Pool}[i] \gets \textit{Pool}[i] \oplus A[i] \oplus B[i]$
        \STATE $\textit{Pool.KV}[i] \gets \textit{KV}[i]$
        \STATE \textbf{if} $\textit{isComplete}(\textit{Pool}[i])$      \textbf{then} $\textit{Pool.deactivate}(i)$
    \ENDFOR
    \STATE $\textit{Window} \gets \texttt{RefillWindow}(\textit{Pool}, W)$
\ENDWHILE
\RETURN $\textit{Pool.sequences}$
\end{algorithmic}
\end{algorithm}

\section{BatchVerify Algorithm Details}
\label{app:batchverify}

Algorithm 3 implements batch verification through a single forward pass over all draft tokens. On the first iteration, the input concatenates the full sequences with their draft tokens ($\mathcal{X} \leftarrow S \oplus D$) to build the initial KV-cache; subsequent iterations process only the draft tokens since prior context is cached. The target model produces logits for all positions, from which we extract predicted tokens via argmax. Mismatch detection is vectorized: we compare predicted tokens against draft tokens and identify the first disagreement position $j$ per sequence. The accepted tokens $A[i]$ consist of all draft tokens up to (but not including) the first mismatch, while the bonus token $B[i]$ is sampled from the target model's distribution at the mismatch position. Critically, this must come from the target model, not the draft model, to preserve output equivalence. Note that because bonus token sampling occurs after the forward pass completes, the bonus token lacks KV-cache entries and must be included in the next forward pass.

An alternative approach, previously employed by vLLM's v0 engine, sidesteps variable acceptance lengths through batch expansion: rather than handling ragged outputs, each sequence is duplicated $K$ times with progressively longer draft prefixes (e.g., $[t_1], [t_1, t_2], [t_1, t_2, t_3]$), all verified in one pass, keeping only the longest correct prefix per sequence. While this maintains rectangular tensors, it incurs $K\times$ memory and compute overhead, resources wasted when early tokens are rejected. This approach was deprecated in vLLM v1 due to GPU memory overflow at scale and incompatibility with CUDA graphs. Our BatchVerify instead accepts the ragged output directly and delegates synchronization to the subsequent unpad-append-repad phase (Section 3.1), avoiding the multiplicative overhead of batch expansion.

\section{Latency-Throughput Tradeoff in Online Serving}
\label{app:latency}

To evaluate performance under realistic online serving conditions, we conducted experiments simulating dynamic request arrivals using full multi-turn conversations from SpecBench. We shuffled conversation turns to maximize length diversity and inserted new turns into the next available batch, measuring both throughput (tokens/s) and request completion latency (s) at P50/P90/P99 percentiles. Table~\ref{tab:latency} compares \oursb against \oursx across batch sizes 1, 2, 4, and 8.

Both methods achieve positive speedups under heterogeneous multi-turn workloads, with \oursx obtaining 2--14\% higher throughput than \oursb through cross-batch grouping of same-length sequences. However, \oursx suffers 1.5--7.7$\times$ worse P90/P99 latencies because delaying early requests until same-length partners arrive causes head-of-line blocking. The penalty peaks at batch size 2 (P99 134s) and recedes as batch size grows (34s at batch size 8), since larger batches sustain the throughput (30.5 to 77.5 tokens/s) needed to drain the request queue. In contrast, \oursb maintains predictable tail latency with P99 under 23 seconds across all batch sizes, making it suitable for latency-sensitive online serving where service-level objectives must be met. These results demonstrate that the choice between \oursb and \oursx depends on deployment requirements: \oursb provides stable, predictable latency for interactive applications, while \oursx maximizes throughput for offline batch processing where individual request latency is less critical. By offering both algorithms with explicit algorithmic correctness guarantees, our work enables practitioners to match their batch speculation strategy to their specific operational constraints. 

\section{Continuous Batching Systems}
\label{app:vllm_sglang}

\begin{tcolorbox}[colback=gray!8, colframe=gray!50, title=Scope]
Continuous-batching systems (vLLM, SGLang) address raggedness through memory layout rather than synchronization: variable-length packing with paged attention stores sequences in a flat buffer, so Invariant~I1 never arises (Family~B, Section~\ref{sec:design-space}). We report two regimes of vLLM: the v0 engine used in our main comparison (Table~\ref{tab:continuous_batching_verification}), and vLLM~v1, which added external-draft speculative decoding in January 2026 (Table~\ref{tab:vllm_v1}). vLLM~v1 is output-equivalent on H100, confirming that the varlen/paged layout is correct where its kernels are available; the divergences reported below reflect numerical precision and feature scope.
\end{tcolorbox}

\paragraph{Architectural differences.} Static batching processes a fixed batch of $B$ sequences together, requiring rectangular tensor alignment (Invariant I1) and facing the ragged tensor synchronization challenge when sequences accept different numbers of draft tokens. Continuous batching systems (e.g., vLLM, SGLang) process requests using variable-length packing, sidestepping I1, but still require scatter-gather across requests and rollback both position-ID and KV-cache for rejected tokens, a one-dimensional analog of alignment. Critically, I2 remains a prerequisite for algorithmic correctness regardless of batching strategy. When we ran our main comparison, neither engine exposed external-draft speculative decoding through continuous batching; vLLM~v1 has since added it (Table~\ref{tab:vllm_v1}).

\paragraph{Systems evaluated.} For informational comparison, we evaluated: \textbf{vLLM}~\citep{kwon2023efficient} (which subsumes TETRIS~\citep{wu2024tetris} and TurboSpec~\citep{liu2025turbospecclosedloopspeculationcontrol} as vLLM forks) and \textbf{SGLang-EAGLE}~\citep{zheng2024sglang,li2024eagle2,li2025eagle3scalinginferenceacceleration}. We used the vLLM v0 engine, the only version supporting external-draft speculative decoding when we ran these experiments; v1 added a varlen replacement in January 2026 (Table~\ref{tab:vllm_v1}). SGLang is compatible only with the EAGLE family; we compare Vicuna-7B/EAGLE2 and Qwen3-8B/EAGLE3 (no available weights for GLM-4). We further tested \textbf{SGLang-EAGLE-Deterministic}~\citep{lmsys_sglang_deterministic_2025}, which enables deterministic execution to reduce numerical variance.


\begin{algorithm}[tb]
\caption{BatchVerify: Single Forward Pass}
\label{alg:batch_verify}
\footnotesize
\begin{algorithmic}[1]
\REQUIRE Target model $\mathcal{M}_t$, Sequences $\mathcal{S}$, Draft tokens $D$, KV cache
\ENSURE Accepted tokens $A$, Bonus tokens $B$
\IF{first iteration}
\STATE $\mathcal{X} \gets \mathcal{S} \oplus D$
\ELSE
\STATE $\mathcal{X} \gets  D$
\ENDIF
\STATE $\textit{logits}, \textit{KVCache} \gets \mathcal{M}_t(\mathcal{X}, \textit{KVCache})$
\STATE $\textit{pred\_tokens} \gets \arg\max(\textit{logits, dim=vocab})$
\STATE \COMMENT{Vectorized first mismatch detection}
\STATE $\textit{matches} \gets (\textit{pred\_tokens} = D)$
\STATE $J \gets \arg\max(\neg\textit{matches, dim=seq})$
\STATE \COMMENT{Ragged shape acceptance, no vectorization}
\FOR{each sequence $i$ in batch}
    \STATE $A[i] \gets D[i][:j]$
    \STATE \COMMENT{Get bonus token from first mismatch}
    \STATE $\textit{bonus\_logit} \gets \textit{logits}[i, |\mathcal{S}[i]| + j]$
    \STATE $B[i] \gets \arg\max(\textit{bonus\_logit})$
\ENDFOR
\RETURN $A, B, KVCache$
\end{algorithmic}
\end{algorithm}

\begin{figure}[h]
    \centering
    \includegraphics[width=0.7\linewidth]{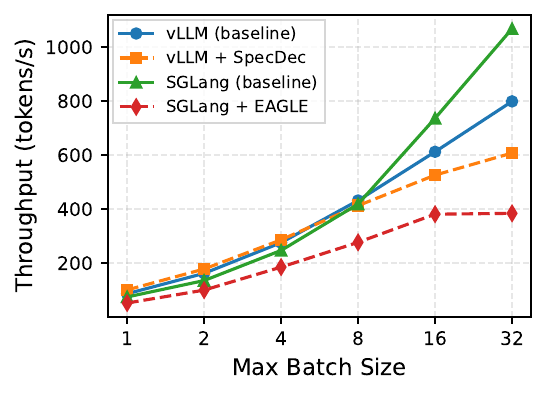}
    \caption{Speculative decoding lags non-speculative baselines, with larger batches further degrading throughput.}
    \label{fig:vllm-scaling}
\end{figure}

\paragraph{Throughput observations.}
Figure~\ref{fig:vllm-scaling} shows throughput characteristics of production frameworks. These systems use continuous batching with dynamically varying effective batch sizes, so the x-axis is a ``maximum batch size'' rather than the fixed batch size of our experiments. The observed patterns (vLLM's speculative decoding underperforming its baseline at high concurrency, SGLang+EAGLE being slower than non-speculative generation) reflect the challenges of integrating speculation with continuous batching. Independent evaluation by \citet{liu2025speculative} confirms these challenges: they report EAGLE speedup degrading from 1.73$\times$ to 1.21$\times$ as batch size scales from 1 to 128, with verification consuming 42--95\% of execution time. vLLM's v0 engine used batch expansion, deprecated in v1 (vllm-project/vllm\#17984) due to $K\times$ memory overhead.

\begin{table*}[t]
\centering
\footnotesize
\begin{tabular}{l|cc|cc|cc}
\toprule
\multirow{2}{*}{\textbf{Method}} & \multicolumn{2}{c|}{\textbf{Vicuna}} & \multicolumn{2}{c|}{\textbf{Qwen3}} & \multicolumn{2}{c}{\textbf{GLM4}} \\
& E & \cellcolor{lightblue}P & E & \cellcolor{lightblue}P & E & \cellcolor{lightblue}P \\
\midrule
vLLM + Spec        & 96.9 & \cellcolor{lightblue}98.0 & 65.6 & \cellcolor{lightblue}78.5 & 72.7 & \cellcolor{lightblue}84.5 \\
SGLang + EAGLE     & 69.8 & \cellcolor{lightblue}79.5 & 47.7 & \cellcolor{lightblue}65.4 & -- & -- \\
SGLang + EAGLE + Det & 85.0 & \cellcolor{lightblue}90.0 & 50.6 & \cellcolor{lightblue}69.5 & -- & -- \\
\bottomrule
\end{tabular}
\caption{Correctness of continuous batching systems. Exact match (E) and partial match (P) scores compared against each system's own non-speculative mode.}
\label{tab:continuous_batching_verification}
\end{table*}

\paragraph{Accuracy observations.} Table~\ref{tab:continuous_batching_verification} shows accuracy metrics for vLLM and SGLang. Both systems are accurate on Vicuna but degrade on Qwen3, suggesting model-specific sensitivities to position encoding or numerical precision. SGLang-EAGLE-Deterministic~\citep{lmsys_sglang_deterministic_2025} helps disambiguate the cause: the improvement from 69.8\% to 85.0\% on Vicuna confirms that most divergences stem from floating-point non-determinism~\citep{he2025nondeterminism} rather than algorithmic bugs. Concurrent work by \citet{liu2025speculative} reaches similar conclusions using output length correlation, though our exact and partial match metrics provide finer-grained verification by distinguishing complete sequence match from partial divergence. Note that SGLang only supports EAGLE-family drafters, not external draft models.

The vLLM~v1 batch-invariant-kernel results are reported in Section~\ref{sec:verification_exp}.


\paragraph{Architectural implications.} The contrast between batch expansion and our approach illustrates two strategies for handling ragged tensors: resource duplication versus explicit synchronization. Batch expansion maintains rectangular tensors by duplicating work, trading $K\times$ memory for implementation simplicity. Our unpad-append-repad approach instead accepts raggedness and pays synchronization costs only when sequences diverge. Integrating this synchronization-based approach with continuous batching and paged attention remains an open problem for production deployment.

\section{Discussion}
\label{app:discussion}
Batch parallelism and per-sequence speculative decoding represent two orthogonal acceleration strategies that should, in principle, multiply together: batching exploits GPU parallelism across sequences while speculation reduces sequential token dependencies within each sequence. However, when combining these techniques in practice, the ragged tensor problem emerges as a fundamental obstacle that breaks this multiplicative relationship. Different sequences accept different numbers of draft tokens, desynchronizing position IDs, attention masks, and KV-cache state across the batch. A correctly implemented batch speculative decoding, as we provide through \oursb and \oursx, preserves the multiplicative gains but incurs a realignment cost that consumes up to 40\% of computation. Our work establishes the synchronization invariants required for algorithmic correctness, quantifies their cost under contiguous layouts, and demonstrates through cross-batch scheduling that intelligent system design can mitigate, though not eliminate, the overhead of maintaining output equivalence at scale.

\section{Reproducibility Statement}
\label{app:reproducibility}

We provide our complete implementation in the Supplementary Material, including all hyperparameters, experimental configurations, and model specifications detailed in Section~\ref{sec:exp}. Our output equivalence verification framework using exact and partial match metrics enables deterministic validation of both our results and future implementations. All experiments use publicly available models and datasets for reproducibility.

\paragraph{Model Configurations.}
We evaluate three target-draft model pairs: Vicuna-7B (lmsys/vicuna-7b-v1.3) with a 68M draft model (double7/vicuna-68m), Qwen3-8B (Qwen/Qwen3-8B) with Qwen3-0.6B (Qwen/Qwen3-0.6B), and GLM-4-9B (zai-org/GLM-4-9B-0414) with a 0.6B draft model (jukofyork/GLM-4.5-DRAFT-0.6B-v3.0). All models were loaded in FP16 precision with greedy decoding (temperature=0, top\_p=1.0) to ensure deterministic outputs. We use five draft tokens per speculation round across all experiments unless otherwise specified.

\paragraph{Production Systems Configuration.}
We evaluated two production inference systems for comparison: vLLM version 0.9.1 and SGLang commit c4e314f (the deterministic decoding had not merged to the  stable version when we ran these experiments, so we compiled it from source). For vLLM, we used the V0 engine with speculative decoding enabled, as the V1 engine  did not yet support draft-model speculative decoding\footnote{\url{https://github.com/vllm-project/vllm/issues/21797}} (added in January 2026; see Table~\ref{tab:vllm_v1}). For SGLang, we used EAGLE-based speculation  with model-specific draft models: yuhuili/EAGLE-Vicuna-7B-v1.3 for Vicuna-7B and Tengyunw/qwen3\_8b\_eagle3 for Qwen3-8B. GLM-4 was not evaluated with SGLang due to the unavailability of compatible EAGLE draft models. We tested both standard SGLang inference and SGLang with deterministic mode enabled~\citep{lmsys_sglang_deterministic_2025} to isolate floating-point non-determinism from algorithmic correctness issues.

\section{Disclose on LLM Usage}
\label{app:llm}

We used Large Language Models as assistive tools in preparing this manuscript: GPT-5 and Claude Opus 4.1 for polishing writing (grammar correction and sentence restructuring), generating conceptual diagram illustrations, and writing unit tests; NVIDIA/Parakeet-TDT-0.6B-v3 for voice input transcription. LLMs were not used for research ideation, experimental design, data analysis, or scientific conclusions. All core algorithmic implementations and scientific contributions are solely from the authors. We take full responsibility for all content in this paper.


\end{document}